\documentclass[runningheads]{llncs}

\usepackage{eccv}
\usepackage{eccvabbrv}
\usepackage{graphicx}
\usepackage{booktabs}
\usepackage{placeins}
\usepackage{float}
\usepackage{tabularx}
\usepackage{array}
\usepackage{microtype}
\usepackage[accsupp]{axessibility}
\usepackage{hyperref}
\usepackage{orcidlink}

\newcommand{\pt}{\hat{\mathbf p}}

\hypersetup{
  colorlinks=true,
  linkcolor=eccvblue,
  citecolor=eccvblue,
  urlcolor=eccvblue,
  pdftitle={Surface-to-Skeleton 3D Cephalometry: Estimating Hidden Skeletal Landmarks from CT-Derived External Soft-Tissue Surfaces},
  pdfauthor={Tomoki Abe, Taiki Kanaya, Kazuki Saita, Mao Noda, Chie Tachiki, Yasushi Nishii, and Hideo Saito},
  pdfkeywords={3D cephalometry, surface-to-skeleton inference, point clouds, hidden landmark estimation}
}

\begin{document}

\title{Surface-to-Skeleton 3D Cephalometry: Estimating Hidden Skeletal Landmarks from CT-Derived External Soft-Tissue Surfaces
} 
\titlerunning{Surface-to-Skeleton 3D Cephalometric Landmark Estimation}
\author{Tomoki Abe\inst{1}\orcidlink{0009-0005-5368-851X} \and
Taiki Kanaya\inst{1}\orcidlink{0009-0009-0554-9514} \and
Kazuki Saita\inst{1}\orcidlink{0009-0002-7280-5695} \and
Mao Noda\inst{2}\orcidlink{0009-0004-6830-6302} \and
Chie Tachiki\inst{2}\orcidlink{0000-0003-0614-9159} \and
Yasushi Nishii\inst{2}\orcidlink{0000-0001-7695-1564} \and
Hideo Saito\inst{1}\orcidlink{0000-0002-2421-9862}}

\authorrunning{T.~Abe et al.}
\institute{Keio University, Yokohama, Japan \and
Tokyo Dental College, Tokyo, Japan \\
\email{abe.tomoki.1106@keio.jp}}

\maketitle

\begin{abstract}
Existing 3D facial-landmark methods localize points on visible skin, but whether CT-defined internal skeletal landmarks can be inferred from external soft-tissue geometry remains unclear. We formulate a coordinate-consistent surface-to-skeleton task using same-acquisition CT-derived surfaces, separating estimation from optical-to-CT registration, scanner-domain, and acquisition-state effects, with coverage analyzed separately. From 240 clinical CT scans from two hospitals, we construct a locked retrospective protocol pairing CT-derived external soft-tissue point clouds with 21 skeletal landmarks
and three visible soft-tissue landmarks. An integrated hierarchical
point-cloud model achieves 2.97~mm mean radial error on skeletal landmarks and 3.03~mm on deep or surface-invisible landmarks in 40 held-out patients. Patient-mismatch controls support patient-specific signal beyond a fixed population configuration or global similarity alone, while coverage ablations indicate dependence on non-anterior geometry. Optical-transfer diagnostics reveal substantial coverage-related and global-configuration components, although deployable optical inference remains unresolved. These results answer the controlled feasibility question affirmatively and provide a basis for hidden skeletal landmark inference.

\end{abstract}

\section{Introduction}
Cephalometric landmarks support orthodontic diagnosis, orthognathic planning, and longitudinal assessment by quantifying the spatial relations among the cranial base, maxilla, mandible, and facial soft tissue \cite{proffit2018contemporary,downs1948,yoshikawa2022}. In current three-dimensional workflows, skeletal landmarks are identified on CT or cone-beam CT (CBCT), either manually or by learning-based systems. Repeated radiographic acquisition is undesirable. Manual landmarking remains time-consuming and observer-dependent \cite{baumrind1971reliability,schlicher2012consistency,hassan2013precision}.

External soft-tissue geometry can instead be acquired by stereophotogrammetry, structured light, or photogrammetry. Existing 3D facial-landmark methods accurately localize points defined on the visible skin surface \cite{berends2024automated,burger2025twostage,qiu2026attention,bao2026fstnet}. The harder question addressed here is different: can the original CT-defined positions of internal skeletal landmarks, including Sella, Basion, Porion, Foramen Ovale, and Condylion Superius, be inferred from same-acquisition CT-derived external soft-tissue geometry? These landmarks have no direct surface correspondence, and similar external surfaces can correspond to different internal configurations. Their locations must therefore be inferred through population-learned relations between soft-tissue shape and craniofacial anatomy, conditioned on each patient's external geometry.

The eventual clinical motivation is to estimate cephalometric landmarks and measurements from independently acquired 3D facial scans, which can be obtained without ionizing radiation. However, a direct optical-to-CT study combines several sources of error: the intrinsic surface-to-skeleton ambiguity, optical-to-CT registration, scanner-domain shift, incomplete surface coverage, facial expression and posture differences, and the time interval between acquisitions. These factors make it difficult to determine whether failure arises because external soft-tissue geometry lacks skeletal information, or because the transfer problem is not yet solved. We therefore isolate the intrinsic feasibility question by using external soft-tissue surfaces segmented from the same CT volumes as the skeletal labels. This coordinate-consistent CT-derived setting asks whether the external 3D head surface itself contains predictive signal for CT-defined landmarks, while real optical-scan deployment remains a separate transfer problem. Accordingly, this study provides a controlled feasibility framework and
evaluation protocol for staged translation rather than evidence for immediate
screening, triage, or follow-up use; CT remains necessary when internal anatomy must be directly visualized.

Automatic 3D cephalometric landmarking from CT/CBCT has been studied using spatial-configuration networks, reinforcement learning, multi-center lightweight models, region-division or coarse-to-fine detectors, geodesic learning, and transformer-based geometric constraints \cite{dot2022automatic,kang2021drl,sahlsten2024deep,tao2023automatic,torosdagli2019deep,lu2023cmfnet}; recent reviews place these methods in the context of direct radiographic landmarking rather than surface-only inference \cite{serafin2023accuracy}. Surface cephalometry on bone meshes also requires radiographic acquisition \cite{tanikawa2025surface}. Conversely, optical-surface studies predict soft-tissue landmarks or surface proxies, not the original internal skeletal coordinates \cite{berends2024automated,qiu2026attention,bao2026fstnet}. Two-dimensional facial photographs have shown that external appearance contains some cephalometric signal \cite{takahashi2023cephalometric,shimamura2024accuracy}, motivating a geometry-based 3D investigation.

Prior work has shown population-level covariation between external facial morphology and internal skeletal shape \cite{young2016facial}. Face--skull and face--bone transformation work predicts dense skeletal or facial shapes using statistical-shape, PCA-based, or learned models for surgical planning and craniofacial modeling \cite{nguyen2020statistical,milojevic2024autoskull,ma2023bidirectional,bao2024facial,zhang2025tcfnet}. A recent arXiv preprint extends this direction with a bidirectional face--skull morphable model \cite{wang2025bfsm}. Landmark-level soft-tissue-to-skeletal prediction has also been explored for Nasion localization from soft-tissue references \cite{baldini2025nasion}. These studies show that paired facial and skeletal geometry can support soft-to-hard tissue inference, but they do not isolate the landmark-level inferability question studied here: whether matched external soft-tissue geometry contains patient-specific signal for clinically defined CT landmarks after removing cross-modality registration and inter-acquisition variation. To our knowledge, prior work has not evaluated patient-level prediction of a multi-landmark set of CT-defined 3D cephalometric coordinates, including deep or surface-invisible landmarks, from same-acquisition external soft-tissue point clouds under a coordinate-consistent protocol. Our novelty lies in this controlled landmark-level task formulation and evaluation, rather than in claiming the first soft-to-hard-tissue relationship. Because we are unaware of a protocol-matched benchmark for this setting, we compare against generic point-cloud encoders, a strong coarse-to-local comparator, and a global-descriptor PCA-ridge statistical-shape baseline under the same split.

We make three contributions. First, we formulate a coordinate-consistent CT-derived surface-to-skeleton task and locked retrospective protocol for predicting CT-defined skeletal landmarks from same-acquisition external soft-tissue point clouds. Second, we show that matched external geometry contributes patient-specific predictive signal beyond a fixed population configuration, low-dimensional shape prior, or global similarity alone, and characterize how that signal varies with anatomical observability and surface coverage. Third, we develop an integrated hierarchical point-cloud model and use optical-transfer diagnostics to identify coverage-related and global-configuration components and unresolved local correspondence limitations.

\section{Data and Preprocessing}
\paragraph{Cohort and Annotations.}
The retrospective cohort comprises 240 patients with dentofacial deformity referred for initial orthognathic planning at Tokyo Dental College Chiba Dental Center and Tokyo Dental College Suidobashi Hospital. Indications included skeletal maxillary protrusion, skeletal mandibular protrusion, skeletal open bite, and facial asymmetry. The study protocol was approved by the Tokyo Dental College Ethics Committee (approval no.~1091), and written informed consent was obtained. All CT scans were clinically indicated, and no additional CT was acquired solely for this study. Congenital anomalies, including cleft lip and palate, were excluded. CT scans were acquired in hard-tissue mode with 0.6- or 1.0-mm slice thickness. During CT acquisition, patients were positioned in maximum intercuspation with lips closed, and upright head posture was stabilized with ear rods; facial expression was not otherwise specified. Data were anonymized before analysis and split by patient into 200 training and 40 locked held-out test cases. The training set was further divided into 160 inner-training and 40 inner-validation cases for model selection. Per-case hospital identifiers, age, sex, and BMI were unavailable after anonymization.

For each CT, the external soft-tissue surface was segmented in Proplan CMF 3.0
(Materialise, Belgium), locally corrected when necessary, exported as an STL
mesh, and cleaned of internal air--tissue interfaces. Twenty-one skeletal landmarks were annotated on the bony CT surface
(S, Na, bilateral Po and Or, Ba, bilateral FO, AZ, ZA, Pt.A, ANS, PNS,
Pt.B, Pog, Me, bilateral Go and CoS), and three soft-tissue landmarks
(Prn, Sn, Pog$'$) on the skin surface. One orthodontist annotated all cases, and an experienced orthodontist reviewed every annotation and resolved ambiguity by consensus; this was adjudication/quality control, not an independent repeated annotation.
Formal protocol-matched intra- and inter-observer reliability was unavailable.
An internal cross-software repeat attempt was not protocol-matched and could
not separate workflow/export effects from landmark-placement variability; it is
therefore reported only as a supplement limitation, not as observer reliability.
It was not used for model development, label revision, case exclusion, or the
reported evaluation. All errors are to finalized consensus coordinates rather than to an
observer-independent anatomical truth. Same-acquisition CT-derived surfaces
isolate the estimation task by avoiding separate optical-to-CT registration;
optical-scan validation requires a separate error decomposition. The locked
protocol denotes a fixed split and analysis plan, not a public data release.

\paragraph{Field of View and Registration.}
Clinical acquisition was heterogeneous because routine surgical-planning CT may cover only the clinically required region rather than a standardized full head. We therefore treat missing superior or inferior facial coverage as field-of-view truncation: 40 cases were superior-only truncated, 28 inferior-only truncated, and 36 truncated at both ends (104/240, 43.3\%). The locked test set contained 25 complete-FOV, 6 superior-only truncated, 4 inferior-only truncated, and 5 both-end truncated cases, enabling an exploratory subgroup analysis. All 24 target landmarks were present within the CT volume in every included case; the FOV strata therefore reflect external soft-tissue coverage only, not missing target labels. Some scans included neck, auricular, or posterior head regions. Each mesh was aligned to a training-case template using geometry renderings from 39 viewpoints, MediaPipe facial keypoints \cite{kartynnik2019facial,lugaresi2019mediapipe}, and point-to-plane ICP in Open3D \cite{zhou2018open3d}; no target cephalometric landmark was used for registration. In the primary pipeline, MediaPipe detections were used only to estimate the
rigid initialization and were not model features, although some are spatially
near visible soft-tissue targets. A supplementary coverage-standardized
diagnostic separately used a MediaPipe-derived face bounding box to define the
sampled region, without passing keypoint coordinates to the predictor. The resulting rigid transform was applied to the external surface, bony surface, and all CT-defined landmark coordinates, preserving same-acquisition correspondence in the registered coordinate system. The registered external surface was randomly sampled to 8,192 points and paired with outward normals. Coordinates were divided by 100~mm for network training and converted back to millimeters for evaluation; coordinates and normals are the only model inputs.

For analysis, landmarks are grouped by surface observability: G0, directly visible soft-tissue points (3); G1, frontal skeletal points with relatively strong surface correlates (8); G2, lateral points affected by variable soft-tissue thickness (4); and G3, deep or surface-invisible points (9). Detailed landmark-level results are provided in the supplementary material.

\section{Controlled Task and Integrated Hierarchical Point-Cloud Model}
\begin{figure}[t]
\centering
{\includegraphics[width=\linewidth]{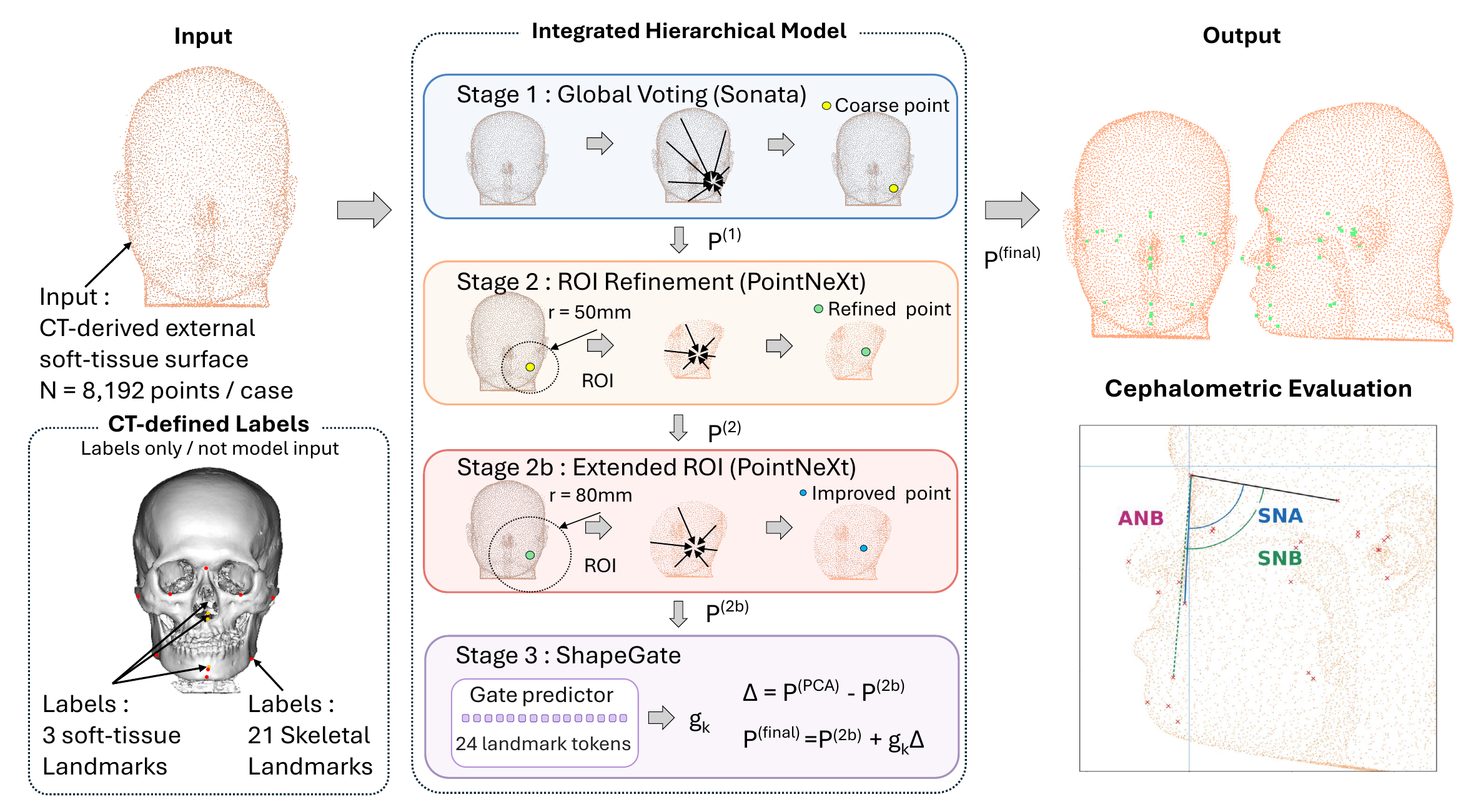}}
\caption{Controlled surface-to-skeleton task and integrated model.
The external soft-tissue point cloud is the only input; CT-defined landmarks are
labels, and the bony surface is not model input. Stages~1--2b perform global and
ROI refinement; ShapeGate applies conservative PCA calibration. Vectra is used
only for secondary transfer diagnostics.}
\label{fig:pipeline}
\end{figure}

The controlled task takes a same-acquisition CT-derived external soft-tissue point cloud as input and predicts 21 skeletal and 3 soft-tissue landmark coordinates in the registered CT frame. Only the external soft-tissue point cloud is provided as input; the bony surface and CT intensities are excluded, and target coordinates are used only as labels. The primary scientific question is whether matched external geometry contains predictive signal for skeletal coordinates; optical-domain transfer is evaluated separately.
Figure~\ref{fig:pipeline} summarizes the integrated four-stage model. It is tailored to the present task through landmark-specific ROI refinement and shape-constrained calibration. Stages~1--2b use the external soft-tissue point cloud and successively produce 24 positions; ShapeGate then operates only on the 24 landmark tokens and their PCA projection.

\subsection{Global Voting and Local Refinement}
\paragraph{Stage 1: Global Voting.}
A Sonata encoder \cite{wu2025sonata}, based on Point Transformer V3 (PTv3) \cite{wu2024ptv3}, processes the entire point cloud. The pre-trained encoder is fully fine-tuned. For each input point $\mathbf x_i$ and landmark $k$, a shared voting head predicts an offset $\hat{\mathbf o}_{ik}$ and confidence logit $\ell_{ik}$. The coarse landmark is the softmax-weighted centroid of proposals,
\begin{equation}
\pt^{(1)}_k =\sum_{i=1}^{N} w_{ik}(\mathbf x_i+\hat{\mathbf o}_{ik}), \quad
\text{ where} \quad w_{ik} =\frac{\exp(\ell_{ik})}{\sum_j\exp(\ell_{jk})}.
\end{equation}
The loss combines Smooth-$\ell_1$ coordinates, proposal supervision, and distance-based confidences, following the voting principle of \cite{qi2019votenet}.

\paragraph{Stage 2: 50-mm ROIs.}
For each landmark, points within a 50-mm sphere centered at $\pt_k^{(1)}$ are represented by relative coordinates, normals, and a landmark-identity code. A PointNeXt encoder \cite{qian2022pointnext} predicts a residual $\hat{\mathbf r}_k$, giving $\pt_k^{(2)}=\pt_k^{(1)}+\alpha_k\hat{\mathbf r}_k$. An inner-validation sweep over $\alpha\in\{0,0.25,0.50,0.75,1.00\}$ selected $\alpha=0.75$ (MRE 3.217~mm) for 23 landmarks; full results are in the supplement. Sella used $\alpha_{\mathrm S}=0$ because the full residual increased its MRE from 4.419 to 4.787~mm. ROIs contain at most 4,096 points; cases with fewer than 128 points retain the Stage-1 estimate.

\paragraph{Stage 2b: Extended ROIs.}
Fourteen high-error landmarks identified on the inner validation set (Pog$'$, S, Na, Ba, AZ, ZA, and bilateral Go, CoS, FO, and Po) receive a second PointNeXt refinement using an 80-mm ROI and up to 8,192 points. Although the 50-mm residual for Sella was suppressed by validation selection, Sella was retained in this broader 80-mm stage because its localization may benefit from wider craniofacial context rather than local surface evidence. Inner-validation search fixed residual damping at 0.4. The other ten landmarks retain their Stage-2 estimates. Stage-1, Stage-2, and Stage-2b models are finally trained on all 200 training cases after these choices are fixed.

\subsection{Conservative Shape Calibration}
Independent local predictions can violate global anatomical consistency. We
therefore fit a PCA shape model \cite{cootes1995active,tothova2018probabilistic}
to the 24-landmark reference configurations of the 160 inner-training cases.
For a Stage-2b prediction, we subtract its landmark centroid, project the
centered configuration into the PCA subspace, reconstruct it, and add back the
original Stage-2b centroid. This yields a centroid-restored PCA projection
$\tilde{\mathbf p}_k$ and shape correction
\begin{equation}
\Delta_k = \tilde{\mathbf p}_k - \pt_k^{(2b)}.
\end{equation}
A shared scalar-output head predicts a case- and landmark-specific gate
$g_k=\sigma(a_k)\in[0,1]$ from the predicted configuration, Stage-1/Stage-2b
disagreement, and landmark identity; its parameters and range, not outputs, are shared.
The final estimate is
\begin{equation}
\pt_k^{\mathrm{final}}
=\pt_k^{(2b)}+g_k\Delta_k.
\end{equation}
The gate does not output an unconstrained XYZ residual, so the final estimate
remains anchored to the Stage-2b prediction and its PCA projection. Full token,
training, and loss details are provided in the supplement. Ablation shows that
this stage acts as modest conservative calibration rather than the main accuracy
driver.

\paragraph{Training.}
Stage~1 is trained for 150 epochs with AdamW \cite{loshchilov2019decoupled}, cosine decay, a $10^{-5}$ encoder learning rate and $10^{-3}$ head learning rate. Stages~2 and 2b are trained for 200 epochs at $10^{-3}$. Augmentation uses small rigid perturbations, Gaussian coordinate noise, point dropout, and ROI-center jitter. ShapeGate converges within five epochs. Architecture and hyperparameters, including the reported gate initialization and loss weights, are selected on the 160/40 inner split; the 40 held-out test cases are not used for fitting or checkpoint selection.

\section{Experiments and Results}
\paragraph{Evaluation.}
Mean radial error (MRE), successful detection rates within 2, 3, and 5~mm (SDR@2/3/5), the 95th-percentile error (P95), and the proportion above 10~mm (Fail@10) are computed over 960 landmark instances. The primary endpoint is all-24-landmark MRE on the locked CT-derived test set; subgroup, degradation, uncertainty, cephalometric, mismatch, and Vectra analyses are exploratory diagnostics. Statistical inference is performed at the patient level: errors are first averaged over the 24 landmarks for each of the 40 test patients. We report 10,000-sample bootstrap confidence intervals and paired permutation tests. All baselines use the same 200/40 split and registered coordinate system. Neural point-cloud baselines use the same point samples and augmentation; the global-descriptor PCA-ridge shape baseline uses fixed descriptors from the same aligned 8,192-point inputs.

We separate three evaluation regimes: the clean CT-derived surface evaluation is the primary coordinate-consistent feasibility test without cross-modality registration; frozen synthetic degradations are stress tests, not scanner simulations; and the secondary Vectra study is an exploratory optical-transfer diagnostic rather than external validation, because optical-to-CT registration, scanner-domain, coverage, and inter-acquisition effects are entangled.

\paragraph{Input Degradation.}
Using the frozen clean model, we changed only input points and normals to probe
coverage, holes/noise, and sparsity. Reference landmarks and model weights were
retained. Exact operators are provided in the supplement.

\subsection{Primary Feasibility and Integrated Model Performance}
On the locked CT-derived test set, the full pipeline achieved 2.97~mm MRE for the 21 skeletal landmarks (SDR@5 88.1\%), including 3.03~mm for the nine deep or surface-invisible landmarks. The prespecified all-24 primary endpoint was 2.91~mm MRE, with a patient-level 95\% CI of 2.73--3.11~mm; the three visible soft-tissue reference landmarks averaged 2.56~mm. These values establish the primary performance for the controlled surface-to-skeleton task. Table~\ref{tab:baseline} compares this performance with fixed population, low-dimensional shape, global point-cloud, and coarse-to-local baselines. PCA-ridge regression from fixed global surface descriptors improves over the training mean but remains well behind learned point-cloud encoders, indicating that a low-dimensional shape model alone is insufficient for the surface-to-skeleton mapping. The full pipeline improves over the Sonata/PTv3 Stage-1 internal baseline by 0.69~mm and over the PTv3 plus PointNeXt 50-mm ROI comparator by 0.22~mm (paired permutation $p<10^{-4}$). SDR@2 is 35.0\%, and gross failures are uncommon (0.7\%), although a 3-mm tolerance is met for only 62.7\% of instances.

Errors are measured against finalized consensus coordinates; Section~\ref{sec:discussion}
discusses protocol-matched observer reliability and interpretation.

\begin{table}[t]
\caption{Held-out test performance on the 40-patient locked test set. Bold indicates the best value and underlining the second-best value within each metric.}
\label{tab:baseline}
\centering
\small
\setlength{\tabcolsep}{3.2pt}
\begin{tabular}{@{}lccccc@{}}
\toprule
Method & MRE$\downarrow$ & SDR@3$\uparrow$ & SDR@5$\uparrow$ & P95$\downarrow$ & Fail@10$\downarrow$\\
 & (mm) & (\%) & (\%) & (mm) & (\%)\\
\midrule
Training-mean landmarks & 8.69 & 8.6 & 22.8 & 17.79 & 31.5\\
Global-desc. PCA-ridge & 6.91 & 11.7 & 35.2 & 13.65 & 18.4\\
PointNet++ \cite{qi2017pointnetplus} & 3.90 & 40.1 & 76.3 & 8.06 & 1.5\\
PointNeXt \cite{qian2022pointnext} & 3.80 & 40.0 & 77.6 & 7.78 & 1.7\\
PointMAE \cite{pang2022masked} & 5.22 & 18.2 & 51.8 & 9.66 & 3.9\\
PTv3 \cite{wu2024ptv3} & 3.79 & 42.8 & 76.8 & 7.90 & 1.9\\
Sonata/PTv3 (Stage 1) & 3.61 & 46.9 & 79.9 & 7.84 & 1.7\\
PTv3 + PointNeXt ROI & \underline{3.14} & \underline{57.7} & \underline{85.5} & \underline{6.72} & \underline{1.4}\\
\textbf{Full pipeline} & \textbf{2.91} & \textbf{62.7} & \textbf{88.0} & \textbf{6.50} & \textbf{0.7}\\
\bottomrule
\end{tabular}
\end{table}

\subsection{Patient-Specificity, Observability, and Model Component Ablation}
To determine whether the primary result reflects matched patient geometry
rather than a learned population template, we first evaluated raw and
similarity-aligned patient-mismatch controls. We used 10 deterministic
non-identity permutations of the 40 held-out final predictions across patients
and evaluated each permuted surface-derived prediction against the original
patient's reference landmarks. Raw mismatch increased overall MRE from 2.91 to
11.87~mm, which is also worse than the training-mean landmark baseline
(8.69~mm). Group-wise errors increased for every observability group, including
G2 lateral landmarks from 3.50 to 11.99~mm and G3 deep landmarks from 3.03 to
11.25~mm (Table~\ref{tab:mismatch}). As an analysis-only control for global
pose and scale confounding, we additionally aligned each mismatched prediction
configuration to the target reference landmarks by a best-fit similarity
transform before computing errors. Because this alignment uses reference
landmarks, it is not a deployable metric; it only tests whether the raw mismatch
result is explained by global similarity differences. The similarity-aligned
mismatch MRE remained 5.36~mm overall, including 5.54~mm for G2 and 4.35~mm for
G3, above the clean matched result. This indicates that matched external
geometry carries patient-specific information beyond a fixed population
configuration or global similarity alone.

\begin{table}[t]
\caption{Frozen patient-mismatch control on the locked test set. Matched is the original patient pairing. Mismatched rows use 10 deterministic non-identity permutations of saved final predictions across patients while keeping each patient's reference landmarks fixed. Similarity alignment uses target reference landmarks and is therefore an analysis-only control, not a deployable metric.}
\label{tab:mismatch}
\centering
\scriptsize
\setlength{\tabcolsep}{1.9pt}
\begin{tabular}{lccccccc}
\toprule
Condition & Pairings & MRE & G0 & G1 & G2 & G3 & Fail@10\\
 & & (mm) & (mm) & (mm) & (mm) & (mm) & (\%)\\
\midrule
Matched & -- & 2.91 & 2.56 & 2.63 & 3.50 & 3.03 & 0.7\\
Mismatched & 10 & 11.87$\pm$0.55 & 12.65$\pm$0.41 & 12.22$\pm$0.55 & 11.99$\pm$0.67 & 11.25$\pm$0.71 & 55.1\\
Mismatch+sim. & 10 & 5.36$\pm$0.16 & 6.36$\pm$0.15 & 6.03$\pm$0.24 & 5.54$\pm$0.21 & 4.35$\pm$0.14 & 6.7\\
\bottomrule
\end{tabular}
\end{table}

Lateral landmarks remain the hardest group (3.50~mm), especially Go L (3.81~mm), AZ (3.42~mm), and ZA (3.39~mm). Their weak geometric signatures and variable soft-tissue thickness limit local inference, but the G2 mismatch increase to 11.99~mm, and to 5.54~mm even after similarity alignment, indicates that these predictions are still patient-specific rather than constant lateral templates. Deep landmarks average 3.03~mm: although invisible, their relative positions are constrained by global craniofacial shape. For this G3 group, MRE decreases from 6.55~mm with the global-descriptor PCA-ridge baseline and 3.91~mm at Stage~1 to 3.22~mm at Stage~2b and 3.03~mm after ShapeGate; together with the G3 mismatch increase to 11.25~mm raw and 4.35~mm after similarity alignment, this indicates that the deep-landmark result is not explained by a simple low-dimensional statistical-shape baseline, fixed population configuration, or global similarity alone. The validation-selected zero residual damping for Sella in the 50-mm Stage~2 refiner is consistent with this pattern: for a deep midline landmark without local surface evidence, the local ROI residual was not trusted beyond the global estimate. Surface Prn and Sn are easiest (1.53 and 1.52~mm). In contrast, soft-tissue Pog$'$ is the single hardest landmark (4.63~mm; 7.5\% above 10~mm), indicating either a poorly defined target, substantial soft-tissue variability, or both. Its 3D location plausibly depends on local chin curvature and acquisition-state-dependent mentalis or lip posture rather than on a uniquely defined surface point.
Several high-error targets have convention-dependent or soft-tissue-dependent
definitions, such as Go as a 3D adaptation of a 2D cephalometric construction. These ambiguities motivate future
protocol-matched observer-reliability studies. Mean gates are largest for PNS, S, and CoS. Because gates are learned without landmark-specific ranges, their magnitudes should be read as correction strengths rather than as direct anatomical observability labels.

ROI refinement drives most of the gain: MRE decreases from 3.61 to 3.09~mm with
50-mm ROIs and to 3.01~mm with the 80-mm Ext14 stage (Table~\ref{tab:ablation}).
Full PCA correction is harmful (3.10~mm); an inner-validation-selected fixed
scalar gate ($g=0.50$) reaches 2.94~mm, and learned gating reaches 2.91~mm.
The patient-level improvement over fixed gating is 0.023~mm (95\% CI
0.011--0.035; paired permutation $p<.001$). Given this small absolute effect,
ShapeGate is conservative configuration calibration, not the main accuracy driver.

\begin{table}[t]
\caption{Ablation (left) and observability-group performance (right). In the ablation subtable, bold indicates the best value within each metric; the group subtable is descriptive and is not ranked by boldface.}
\label{tab:ablation}
\centering
\small
\begin{minipage}[t]{0.53\linewidth}
\centering
\begin{tabular}{lccc}
\toprule
Config. & MRE & SDR@3 & SDR@5\\
 & (mm) & (\%) & (\%)\\
\midrule
Stage 1 & 3.61 & 46.9 & 79.9\\
+ 50-mm ROI & 3.09 & 58.5 & 85.4\\
+ 80-mm Ext14 & 3.01 & 60.6 & 86.7\\
+ free-XYZ residual & 3.00 & 61.7 & 86.7\\
+ PCA corr. ($g=1$) & 3.10 & 57.6 & 86.7\\
+ fixed gate ($g=.50$) & 2.94 & 61.9 & 87.7\\
+ learned gate $[0,1]$ & \textbf{2.91} & \textbf{62.7} & \textbf{88.0}\\
\bottomrule
\end{tabular}
\end{minipage}\hfill
\begin{minipage}[t]{0.43\linewidth}
\centering
\begin{tabular}{lccc}
\toprule
Group & $n$ & MRE & SDR@3\\
 & & (mm) & (\%)\\
\midrule
G0 Surface & 3 & 2.56 & 78.3\\
G1 Frontal & 8 & 2.63 & 68.1\\
G2 Lateral & 4 & 3.50 & 48.8\\
G3 Deep & 9 & 3.03 & 58.9\\
\bottomrule
\end{tabular}
\end{minipage}
\end{table}

Representative best, median, and worst qualitative projections are provided in the supplementary material.

\subsection{Field-of-View Truncation Subgroup}
Because clinical planning CT may omit superior or inferior facial regions, we stratified the locked test cases by field-of-view completeness after the final evaluation. Complete-FOV cases achieved 2.83~mm MRE; superior, inferior, and both-end truncation subgroups achieved 3.20, 2.86, and 3.03~mm, respectively, with Fail@10 below 1.5\% in all strata. These small subgroups are not powered for formal significance testing; they indicate limited degradation under the naturally occurring truncations present in the test set rather than guaranteed robustness to arbitrary missing anatomy. Full subgroup metrics are provided in the supplement.

\subsection{All-Case Anterior-Coverage Ablation}
Because some CT-derived inputs include posterior, auricular, or inferior regions that are not consistently available in front-facing optical scans, we applied frozen analysis-only coverage masks to all 40 test cases. These masks change only the input points and normals, use synthetic percentile thresholds rather than scanner geometry, and are not deployable preprocessing. Mild frontal cropping increased MRE by 0.30~mm (Table~\ref{tab:coverage_ablation}). A stronger face-only crop increased MRE to 3.74~mm, while a tight face-only crop increased it to 4.77~mm and reduced SDR@5 to 64.6\%. G2 lateral and G3 deep errors also increased, especially under the tight crop. Thus, the clean CT-derived result is not a pure anterior-face result: posterior, lateral, and inferior coverage can contribute to the hidden-landmark signal, even though the degradation remains far smaller than the Vectra transfer gap.

\begin{table}[t]
\caption{Frozen all-40 anterior-coverage ablation. Crops are analysis-only synthetic masks and are not scanner simulations; $\Delta$ is relative to clean.}
\label{tab:coverage_ablation}
\centering
\small
\setlength{\tabcolsep}{3.2pt}
\begin{tabular}{lcccccc}
\toprule
Input & MRE & $\Delta$ & G2 & G3 & SDR@5 & Fail@10\\
 & (mm) & (mm) & (mm) & (mm) & (\%) & (\%)\\
\midrule
Clean CT-derived & 2.91 & -- & 3.50 & 3.03 & 88.0 & 0.7\\
Frontal crop & 3.21 & +0.30 & 3.65 & 3.22 & 83.8 & 1.0\\
Face-only crop & 3.74 & +0.82 & 4.07 & 3.43 & 76.2 & 3.3\\
Tight face-only crop & 4.77 & +1.86 & 4.66 & 4.76 & 64.6 & 8.3\\
\bottomrule
\end{tabular}
\end{table}

\subsection{Uncertainty and Abstention}
Frozen internal signals ranked high-error cases without an additional test-set
classifier. ShapeGate correction and PCA-delta magnitudes each achieved AUROC
about 0.80, and retaining the lowest-disagreement 80\% of cases reduced mean
case MRE from 2.91 to 2.74~mm. This exploratory analysis suggests that
correction and disagreement signals may support future abstention or
manual-review prioritization; full metrics are provided in the supplement.

\subsection{Sensitivity to Optical-Like Input Degradation}
Frozen stress tests showed moderate sensitivity to holes/noise and combined
crop, noise, and sparsity perturbations, with combined degradation reaching
4.60~mm final MRE. A sparse-input sweep revealed a density threshold:
4,096 unique points degraded MRE to 3.69~mm, whereas 2,048 points increased it
to 6.61~mm and 1,024 points caused near-collapse (17.64~mm). Downstream stages
still helped but could not recover geometry lost before Stage~1. These operators
omit structured facial changes; full metrics are in the supplement.

\subsection{Cephalometric Agreement}
Measurements were computed identically from predicted and reference landmarks using the respective Po/Or-defined Frankfort Horizontal and sagittal-plane projection. ANB MAE was $1.22^{\circ}$ (P95 $2.95^{\circ}$; Table~\ref{tab:clinical}); SNA/SNB were less accurate because both depend on deep Sella. With Class III defined as ANB $\leq 0^{\circ}$, Class I as $0^{\circ} < \mathrm{ANB} < 4^{\circ}$, and Class II as ANB $\geq 4^{\circ}$, sagittal-class agreement was 34/40 (85.0\%, Wilson 95\% CI 70.9--92.9, $\kappa=0.76$). Agreement was 30/31 among cases at least $1^{\circ}$ from an ANB class boundary but 4/9 among borderline cases. Errors were I$\to$III (3), III$\to$I (1), I$\to$II (1), and II$\to$I (1), with no II$\leftrightarrow$III reversal; this small cohort does not establish clinical safety.

\begin{table}[t]
\caption{Clinical errors (left) and sagittal-class confusion matrix (right; reference rows, predicted columns).}
\label{tab:clinical}
\centering
\small
\renewcommand{\arraystretch}{1.04}
\begin{minipage}[t]{0.49\linewidth}
\centering
\begin{tabular}{lcc}
\toprule
Measurement & MAE & P95\\
\midrule
ANB ($^{\circ}$) & 1.22 & 2.95\\
SNA ($^{\circ}$) & 3.45 & 8.05\\
SNB ($^{\circ}$) & 2.92 & 7.76\\
Mandibular plane--FH ($^{\circ}$) & 1.53 & 4.99\\
\bottomrule
\end{tabular}
\end{minipage}\hfill
\begin{minipage}[t]{0.47\linewidth}
\centering
\begin{tabular}{lccc}
\toprule
Ref. $\backslash$ Pred. & I & II & III\\
\midrule
Class I   & 6 & 1 & 3\\
Class II  & 1 & 9 & 0\\
Class III & 1 & 0 & 19\\
\bottomrule
\end{tabular}
\end{minipage}
\end{table}

\subsection{Optical Transfer: Coverage, Global Configuration, and Remaining Local Gap}
As a secondary exploratory diagnostic, not external validation, we evaluated
14 Vectra scans from eight locked-test patients. These scans were acquired on
preoperative days separate from CT, without CT-style ear-rod fixation, and
therefore differed in scanner domain, surface coverage, head configuration, and
acquisition state. The paired CT-derived landmarks were used only as evaluation
targets, so apparent optical-to-CT error mixes model error, residual
registration error, scanner-domain shift, and non-rigid inter-acquisition
differences.

We performed two complementary diagnostics (Table~\ref{tab:vectra_decomp}).
First, we evaluated frozen full-pipeline transfer on the eight linked patients.
Clean CT-derived inputs yielded a patient-level MRE of 2.79~mm. For each
Vectra scan, we also generated a corresponding CT-derived crop using that
scan's coverage mask; therefore, patients with repeated Vectra acquisitions
could have slightly different paired CT-derived crops. This Vectra-coverage
crop increased patient-level MRE to 3.79~mm, whereas direct frozen Vectra input
yielded 10.96~mm. Thus, reduced coverage explains part, but not most, of the
observed transfer gap. The corresponding displacement from the clean CT-derived
prediction was 2.13~mm for the coverage crop and 10.27~mm for Vectra; this
shift is a diagnostic distance between predictions, not an additive component
of MRE.

Second, because local ROI refinement did not transfer reliably, we isolated
Stage~1 to examine potentially correctable transfer components across all
14 Vectra scans. Without using Vectra scans for training or model selection,
CT-only coverage-standardized training reduced scan-level Stage-1 MRE from
12.40 to 7.94~mm. Analysis-only similarity alignment using reference landmarks
further reduced the coverage-standardized Stage-1 error to 4.07~mm, indicating
a large non-deployable global similarity component that includes residual pose
and registration differences. This aligned residual should not be interpreted
as pure local-correspondence error, because reference-based fitting can absorb
multiple global prediction components. The CT-trained local ROI stage worsened
the coverage-standardized Vectra result from 7.94 to 8.45~mm, indicating
unresolved local optical-domain correspondence. The patient-level full-pipeline
values and scan-level Stage-1 diagnostics use different aggregation schemes and
should not be read as one sequential ablation.

\begin{table}[t]
\caption{Optical-transfer diagnostics. Panel A uses patient-level frozen
full-pipeline summaries on the eight linked locked-test patients and averages
over available scans within patient. For each Vectra scan, the corresponding
CT-derived crop uses that scan's coverage mask. Displacement from clean
prediction measures distance between two predictions and is not an additive
component of MRE. Repeat-scan shift is descriptive within-patient variability
across available repeated scan inputs. Panel B uses scan-level Stage-1
diagnostics over all 14 Vectra scans. Similarity alignment is analysis-only and
uses reference landmarks.}
\label{tab:vectra_decomp}
\centering
\small
\setlength{\tabcolsep}{4.0pt}
\renewcommand{\arraystretch}{1.08}
\begin{tabular*}{0.96\linewidth}{@{\extracolsep{\fill}}lcccc@{}}
\toprule
\multicolumn{5}{l}{\textbf{A. Frozen full-pipeline input comparison}}\\
\midrule
Input & MRE & $\Delta$ vs. clean & Disp. & Repeat shift\\
 & (mm) & (mm) & (mm) & (mm)\\
\midrule
Clean CT-derived & 2.79 & -- & -- & --\\
CT-derived coverage crop & 3.79 & +1.00 & 2.13 & 1.14\\
Frozen full-pipeline Vectra & 10.96 & +8.17 & 10.27 & 3.67\\
\addlinespace[6pt]
\multicolumn{5}{l}{\textbf{B. Stage-1 transfer diagnostics on 14 Vectra scans}}\\
\midrule
Condition & Raw MRE & SDR@5 & Fail@10 & Sim. MRE\\
 & (mm) & (\%) & (\%) & (mm)\\
\midrule
Standard CT training, Stage~1 & 12.40 & 7.1 & 55.4 & 7.53\\
Coverage-std. Stage~1 & 7.94 & 28.9 & 28.3 & 4.07\\
Coverage-std. Stage~2 & 8.45 & 27.4 & 31.2 & 4.28\\
\bottomrule
\end{tabular*}
\end{table}

\section{Discussion}
\label{sec:discussion}
Anterior-coverage ablations qualify the controlled result: predictive signal is not confined to the visible frontal face and can depend on posterior, lateral, and inferior context. Within the integrated model, the 50-mm ROI stage provided the largest architectural gain, while gated PCA correction acted only as a modest conservative calibration. Monotonic improvements across stages under the frozen CT-derived degradation tests further suggest that global voting, local evidence, and shape constraints remain complementary beyond clean inputs.

The cephalometric analysis is encouraging but not yet clinically sufficient.
ANB showed relatively low error and 34/40 sagittal-class agreement, although
shared Sella error can cancel in ANB and does not imply equally accurate
absolute jaw position. Agreement depends on thresholds and cohort composition,
and the three Class-I-to-III errors remain consequential because even
screening-oriented use could misdirect triage without manual review or
abstention. These secondary analyses motivate screening-oriented research but
do not establish clinical screening safety or replacement of CT-based
diagnosis. At the landmark level, reported errors are distances to finalized
consensus coordinates, not to an error-free anatomical truth. Prior 3D-CBCT
studies report landmark-dependent manual variability from sub-millimeter to a
few millimeters under their respective protocols
\cite{schlicher2012consistency,hassan2013precision}. These studies are not protocol-matched to the present annotation workflow, so
the reported errors should not be interpreted as evidence of human-level
performance or observer-independent anatomical accuracy.
Formal protocol-matched observer-reliability estimation remains necessary,
especially for convention-dependent or anatomically ambiguous landmarks.

Optical translation remains a separate problem, but the Vectra pilot refines rather than simply negates the translation story. Anterior-coverage ablations show that stricter face-only masks degrade performance, so the clean CT-derived result should not be interpreted as anterior-face-only performance. Coverage-standardized CT-only retraining produced a large raw Stage-1 Vectra improvement, and analysis-only landmark alignment exposed a large global-configuration component, predominantly rigid rather than scale-related. Nevertheless, Vectra scan inputs remained far from clean CT performance, the local ROI stage was not reliable, and a substantial aligned residual remained. Thus, coverage standardization and deployable global-configuration correction are important next targets but are unlikely to be sufficient alone; scanner-domain, acquisition-state, and non-rigid surface-correspondence effects remain entangled \cite{heike2010stereophotogrammetry,ozsoy2015position,tangthaweesuk2025accuracy}.

The near-term role of the present protocol is therefore methodological rather than clinical. It provides a coordinate-consistent reference for a staged translation pathway: standardize anterior coverage and head posture on CT-derived surfaces, acquire paired CT and optical scans in close temporal proximity under controlled expression and occlusion, quantify registration and surface correspondence error separately, and then perform optical-domain adaptation with explicit abstention. Only after these steps should screening or follow-up utility be evaluated prospectively.

Both hospitals belong to one institution; the pooled split is neither site-held-out
nor external validation. Missing per-case site, scanner, age, sex, and BMI labels
precluded center-held-out and subgroup analyses, so transportability is unestablished.
Other limitations are unavailable protocol-matched observer reliability, the 40-case
orthognathic test cohort with small FOV strata, inner-split final-gate selection,
and four separately trained modules. Broader cohorts and joint/distilled models are needed.

\section{Conclusion}
We evaluated whether CT-defined skeletal landmarks invisible on the face can be inferred from same-acquisition CT-derived external soft-tissue geometry. The integrated hierarchical point-cloud model achieved 2.97~mm MRE on skeletal landmarks and 3.03~mm on deep or surface-invisible landmarks. Patient-mismatch controls show that this performance depends on matched patient geometry rather than a fixed population configuration or global similarity alone, answering the controlled feasibility question affirmatively. Optical-transfer diagnostics further show that the deployment gap is not monolithic: coverage-standardized CT-only training substantially improves global transfer, and analysis-only reference-based alignment reveals a large global-configuration component, while deployable local correspondence remains unresolved. Together, the task, protocol, and analyses establish a controlled basis for hidden skeletal landmark inference.

\section*{Acknowledgements}
This work was partially supported by JSPS KAKENHI Grant Number 23K28112 and 24K13203.

\bibliographystyle{splncs04}
\bibliography{main}

@article{baumrind1971reliability,
  author    = {Baumrind, Sheldon and Frantz, Richard C.},
  title     = {The reliability of head film measurements: 1. Landmark identification},
  journal   = {American Journal of Orthodontics},
  volume    = {60},
  number    = {2},
  pages     = {111--127},
  year      = {1971},
  doi       = {10.1016/0002-9416(71)90028-5}
}

@article{berends2024automated,
  author={Berends, Bo and Bielevelt, Freek and Schreurs, Ruud and Vinayahalingam, Shankeeth and Maal, Thomas and de Jong, Guido},
  title={Fully automated landmarking and facial segmentation on {3D} photographs},
  journal={Scientific Reports}, volume={14}, pages={6463}, year={2024}, doi={10.1038/s41598-024-56956-9}
}

@article{burger2025twostage,
  author={Burger, Jacopo and Blandano, Giorgio and Facchi, Giuseppe Maurizio and Lanzarotti, Raffaella},
  title={{2S-SGCN}: A two-stage stratified graph convolutional network model for facial landmark detection on {3D} data},
  journal={Computer Vision and Image Understanding}, volume={250}, pages={104227}, year={2025}, doi={10.1016/j.cviu.2024.104227}
}

@article{bao2024facial,
  author={Bao, J. and Zhang, X. and Xiang, S. and Liu, H. and Cheng, M. and Yang, Y. and Huang, X. and Xiang, W. and Cui, W. and Lai, H. C. and Huang, S. and Wang, Y. and Qian, D. and Yu, H.},
  title={Deep Learning-Based Facial and Skeletal Transformations for Surgical Planning},
  journal={Journal of Dental Research}, volume={103}, number={8}, pages={809--819}, year={2024}, doi={10.1177/00220345241253186}
}

@article{bao2026fstnet,
  author    = {Bao, Han and He, Zhidong and Wu, Jiasong and Baxter, John and Senhadji, Lotfi and Shu, Huazhong and Liu, Luwei and Yan, Bin},
  title     = {A new automated {3D} facial soft tissue landmarking method via deep learning},
  journal   = {Journal of Dentistry},
  volume    = {173},
  pages     = {106742},
  year      = {2026},
  doi       = {10.1016/j.jdent.2026.106742}
}

@article{ma2023bidirectional,
  author={Ma, Lei and Lian, Chunfeng and Kim, Daeseung and Xiao, Deqiang and Wei, Dongming and Liu, Qin and Kuang, Tianshu and Ghanbari, Maryam and Li, Guoshi and Gateno, Jaime and Shen, Steve G. F. and Wang, Li and Shen, Dinggang and Xia, James J. and Yap, Pew-Thian},
  title={Bidirectional prediction of facial and bony shapes for orthognathic surgical planning},
  journal={Medical Image Analysis}, volume={83}, pages={102644}, year={2023}, doi={10.1016/j.media.2022.102644}
}

@article{nguyen2020statistical,
  author={Nguyen, Tan-Nhu and Tran, Vi-Do and Nguyen, Ho-Quang and Dao, Tien-Tuan},
  title={A statistical shape modeling approach for predicting subject-specific human skull from head surface},
  journal={Medical \& Biological Engineering \& Computing},
  volume={58},
  number={10},
  pages={2355--2373},
  year={2020},
  doi={10.1007/s11517-020-02219-4}
}

@inproceedings{milojevic2024autoskull,
  author    = {Milojevic, Aleksandar and Peter, Daniel and Huber, Niko B. and Azevedo, Luis and Latyshev, Andrei and Sailer, Irena and Gross, Markus and Thomaszewski, Bernhard and Solenthaler, Barbara and G{\"o}zc{\"u}, Baran},
  title     = {{AutoSkull}: Learning-based skull estimation for automated pipelines},
  booktitle = {Medical Image Computing and Computer Assisted Intervention -- {MICCAI} 2024},
  series    = {Lecture Notes in Computer Science},
  volume    = {15007},
  pages     = {109--118},
  year      = {2024},
  publisher = {Springer Nature Switzerland},
  doi       = {10.1007/978-3-031-72104-5_11}
}

@inproceedings{baldini2025nasion,
  author    = {Baldini, Benedetta and Yazdi, Ali Shadman and Serafin, Marco and Tartaglia, Gianluca M. and Baselli, Giuseppe},
  title     = {Predicting skeletal landmarks from soft-tissue landmarks using machine learning: A study on Nasion localization},
  booktitle = {2025 47th Annual International Conference of the {IEEE} Engineering in Medicine and Biology Society ({EMBC})},
  pages     = {1--5},
  year      = {2025},
  publisher = {IEEE},
  doi       = {10.1109/EMBC58623.2025.11251704}
}

@article{cootes1995active,
  author    = {Cootes, Timothy F. and Taylor, Christopher J. and Cooper, David H. and Graham, Jim},
  title     = {Active shape models---their training and application},
  journal   = {Computer Vision and Image Understanding},
  volume    = {61},
  number    = {1},
  pages     = {38--59},
  year      = {1995},
  doi       = {10.1006/cviu.1995.1004}
}

@article{downs1948,
  author    = {Downs, William B.},
  title     = {Variations in facial relationships: Their significance in treatment and prognosis},
  journal   = {American Journal of Orthodontics},
  volume    = {34},
  number    = {10},
  pages     = {812--840},
  year      = {1948},
  doi       = {10.1016/0002-9416(48)90015-3}
}

@article{efron1979bootstrap,
  author={Efron, Bradley},
  title={Bootstrap methods: Another look at the jackknife},
  journal={The Annals of Statistics},
  volume={7},
  number={1},
  pages={1--26},
  year={1979},
  doi={10.1214/aos/1176344552}
}

@book{good2005permutation,
  author={Good, Phillip},
  title={Permutation, Parametric, and Bootstrap Tests of Hypotheses},
  edition={3},
  series={Springer Series in Statistics},
  publisher={Springer},
  address={New York},
  year={2005},
  doi={10.1007/b138696}
}

@article{kang2021drl,
  author={Kang, Sung Ho and Jeon, Kiwan and Kang, Sang-Hoon and Lee, Sang-Hwy},
  title={{3D} cephalometric landmark detection by multiple stage deep reinforcement learning},
  journal={Scientific Reports}, volume={11}, pages={17509}, year={2021}, doi={10.1038/s41598-021-97116-7}
}

@article{dot2022automatic,
  author={Dot, Gauthier and Schouman, Thomas and Chang, Shih and Rafflenbeul, Fr{\'e}d{\'e}ric and Kerbrat, Aur{\'e}lie and Rouch, Philippe and Gajny, Laurent},
  title={Automatic {3-Dimensional} Cephalometric Landmarking via Deep Learning},
  journal={Journal of Dental Research}, volume={101}, number={11}, pages={1380--1387}, year={2022}, doi={10.1177/00220345221112333}
}

@inproceedings{kartynnik2019facial,
  author    = {Kartynnik, Yury and Ablavatski, Artsiom and Grishchenko, Ivan and Grundmann, Matthias},
  title     = {Real-time facial surface geometry from monocular video on mobile {GPUs}},
  booktitle = {{CVPR} Workshop on Computer Vision for Augmented and Virtual Reality},
  address   = {Long Beach, CA},
  year      = {2019},
  publisher = {IEEE},
  note      = {arXiv preprint arXiv:1907.06724},
  doi       = {10.48550/arXiv.1907.06724},
  eprint    = {1907.06724},
  archivePrefix = {arXiv}
}

@inproceedings{loshchilov2019decoupled,
  author    = {Loshchilov, Ilya and Hutter, Frank},
  title     = {Decoupled weight decay regularization},
  booktitle = {International Conference on Learning Representations ({ICLR})},
  year      = {2019}
}

@article{lu2023cmfnet,
  author={Lu, Gang and Shu, Huazhong and Bao, Han and Kong, Youyong and Zhang, Chen and Yan, Bin and Zhang, Yuanxiu and Coatrieux, Jean-Louis},
  title={{CMF-Net}: Craniomaxillofacial landmark localization on {CBCT} images using geometric constraint and transformer},
  journal={Physics in Medicine \& Biology}, volume={68}, number={9}, pages={095020}, year={2023}, doi={10.1088/1361-6560/acb483}
}

@misc{lugaresi2019mediapipe,
  author    = {Lugaresi, Camillo and Tang, Jiuqiang and Nash, Hadon and McClanahan, Chris and Uboweja, Esha and Hays, Michael and Zhang, Fan and Chang, Chuo-Ling and Yong, Ming Guang and Lee, Juhyun and Chang, Wan-Teh and Hua, Wei and Georg, Manfred and Grundmann, Matthias},
  title     = {{MediaPipe}: A framework for building perception pipelines},
  year      = {2019},
  note      = {arXiv preprint arXiv:1906.08172},
  doi       = {10.48550/arXiv.1906.08172},
  eprint    = {1906.08172},
  archivePrefix = {arXiv}
}

@inproceedings{pang2022masked,
  author    = {Pang, Yatian and Wang, Wenxiao and Tay, Francis E. H. and Liu, Wei and Tian, Yonghong and Yuan, Li},
  title     = {Masked autoencoders for point cloud self-supervised learning},
  booktitle = {Computer Vision -- {ECCV} 2022},
  series    = {Lecture Notes in Computer Science},
  volume    = {13662},
  pages     = {604--621},
  year      = {2022},
  publisher = {Springer},
  doi       = {10.1007/978-3-031-20086-1_35}
}

@book{proffit2018contemporary,
  author    = {Proffit, William R. and Fields, Henry W. and Larson, Brent E. and Sarver, David M.},
  title     = {Contemporary Orthodontics},
  edition   = {6},
  publisher = {Elsevier},
  address   = {Philadelphia, PA},
  year      = {2018},
  isbn      = {978-0-323-54387-3}
}

@inproceedings{qi2017pointnetplus,
  author    = {Qi, Charles R. and Yi, Li and Su, Hao and Guibas, Leonidas J.},
  title     = {{PointNet++}: Deep hierarchical feature learning on point sets in a metric space},
  booktitle = {Advances in Neural Information Processing Systems 30},
  year      = {2017}
}

@inproceedings{qi2019votenet,
  author    = {Qi, Charles R. and Litany, Or and He, Kaiming and Guibas, Leonidas J.},
  title     = {Deep {Hough} voting for {3D} object detection in point clouds},
  booktitle = {Proceedings of the {IEEE}/{CVF} International Conference on Computer Vision ({ICCV})},
  pages     = {9277--9286},
  year      = {2019},
  publisher = {IEEE}
}

@inproceedings{qian2022pointnext,
  author    = {Qian, Guocheng and Li, Yuchen and Peng, Houwen and Mai, Jinjie and Hammoud, Hasan and Elhoseiny, Mohamed and Ghanem, Bernard},
  title     = {{PointNeXt}: Revisiting {PointNet++} with improved training and scaling strategies},
  booktitle = {Advances in Neural Information Processing Systems 35},
  pages     = {23192--23204},
  year      = {2022}
}

@article{qiu2026attention,
  author={Qiu, Tao and Hu, Chaoran and Zhang, Jingyu and Wu, Fuli and Wang, Huiming and Liu, Xiangtao and Sun, Mouyuan},
  title={{Attention-Base} deep learning for {3D} craniofacial soft tissue landmark detection and diagnosis in orthodontics},
  journal={Scientific Reports}, volume={16}, pages={729}, year={2026}, doi={10.1038/s41598-025-30383-w}
}

@article{sahlsten2024deep,
  author    = {Sahlsten, Jaakko and J{\"a}rnstedt, Jorma and Jaskari, Joel and Naukkarinen, Hanna and Mahasantipiya, Phattaranant and Charuakkra, Arnon and Vasankari, Krista and Hietanen, Ari and Sundqvist, Osku and Lehtinen, Antti and Kaski, Kimmo},
  title     = {Deep learning for {3D} cephalometric landmarking with heterogeneous multi-center {CBCT} dataset},
  journal   = {PLOS ONE},
  volume    = {19},
  number    = {6},
  pages     = {e0305947},
  year      = {2024},
  doi       = {10.1371/journal.pone.0305947}
}

@article{tao2023automatic,
  author={Tao, Leran and Li, Meng and Zhang, Xu and Cheng, Mengjia and Yang, Yang and Fu, Yijiao and Zhang, Rongbin and Qian, Dahong and Yu, Hongbo},
  title={Automatic craniomaxillofacial landmarks detection in {CT} images of individuals with dentomaxillofacial deformities by a two-stage deep learning model},
  journal={BMC Oral Health}, volume={23}, pages={876}, year={2023}, doi={10.1186/s12903-023-03446-5}
}

@article{takahashi2023cephalometric,
  author={Takahashi, Kaisei and Shimamura, Yui and Tachiki, Chie and Nishii, Yasushi and Hagiwara, Masafumi},
  title={Cephalometric landmark detection without {X-rays} combining coordinate regression and heatmap regression},
  journal={Scientific Reports}, volume={13}, pages={20011}, year={2023}, doi={10.1038/s41598-023-46919-x}
}

@article{shimamura2024accuracy,
  author={Shimamura, Yui and Tachiki, Chie and Takahashi, Kaisei and Matsunaga, Satoru and Takaki, Takashi and Hagiwara, Masafumi and Nishii, Yasushi},
  title={Accuracy of cephalometric landmark and cephalometric analysis from lateral facial photograph by using {CNN}-based algorithm},
  journal={Scientific Reports}, volume={14}, pages={31089}, year={2024}, doi={10.1038/s41598-024-82230-z}
}

@article{tanikawa2025surface,
  author={Tanikawa, Chihiro and Nakamura, Hiroyuki and Mimura, Takaaki and Uemura, Yume and Yamashiro, Takashi},
  title={Development of artificial intelligence-supported automatic three-dimensional surface cephalometry},
  journal={Orthodontics \& Craniofacial Research}, volume={28}, number={4}, pages={636--646}, year={2025}, doi={10.1111/ocr.12914}
}

@article{torosdagli2019deep,
  author={Torosdagli, Neslisah and Liberton, Denise K. and Verma, Payal and Sincan, Murat and Lee, Janice S. and Bagci, Ulas},
  title={Deep geodesic learning for segmentation and anatomical landmarking},
  journal={IEEE Transactions on Medical Imaging}, volume={38}, number={4}, pages={919--931}, year={2019}, doi={10.1109/TMI.2018.2875814}
}

@inproceedings{tothova2018probabilistic,
  author    = {T{\'o}thov{\'a}, Katar{\'i}na and Parisot, Sarah and Lee, Matthew and Puyol-Ant{\'o}n, Esther and King, Andrew P. and Pollefeys, Marc and Konukoglu, Ender},
  title     = {Probabilistic {3D} surface reconstruction from sparse {MRI} information},
  booktitle = {Medical Image Computing and Computer Assisted Intervention -- {MICCAI} 2020},
  series    = {Lecture Notes in Computer Science},
  volume    = {12261},
  pages     = {813--823},
  year      = {2020},
  publisher = {Springer International Publishing},
  doi       = {10.1007/978-3-030-59710-8_79}
}

@inproceedings{wu2024ptv3,
  author    = {Wu, Xiaoyang and Jiang, Li and Wang, Peng-Shuai and Liu, Zhijian and Liu, Xihui and Qiao, Yu and Ouyang, Wanli and He, Tong and Zhao, Hengshuang},
  title     = {{Point Transformer V3}: Simpler, faster, stronger},
  booktitle = {Proceedings of the {IEEE}/{CVF} Conference on Computer Vision and Pattern Recognition ({CVPR})},
  pages     = {4840--4851},
  year      = {2024},
  publisher = {IEEE}
}

@inproceedings{wu2025sonata,
  author    = {Wu, Xiaoyang and DeTone, Daniel and Frost, Duncan and Shen, Tianwei and Xie, Chris and Yang, Nan and Engel, Jakob and Newcombe, Richard and Zhao, Hengshuang and Straub, Julian},
  title     = {{Sonata}: Self-supervised learning of reliable point representations},
  booktitle = {Proceedings of the {IEEE}/{CVF} Conference on Computer Vision and Pattern Recognition ({CVPR})},
  pages     = {22193--22204},
  year      = {2025},
  publisher = {IEEE}
}

@article{yoshikawa2022,
  author={Yoshikawa, Hiroshi and Tanikawa, Chihiro and Ito, Shinsuke and Tsukiboshi, Yosuke and Ishii, Hitomi and Kanomi, Ryuzo and Yamashiro, Takashi},
  title={A three-dimensional cephalometric analysis of {Japanese} adults and its usefulness in orthognathic surgery: A retrospective study},
  journal={Journal of Cranio-Maxillofacial Surgery}, volume={50}, number={4}, pages={353--363}, year={2022}, doi={10.1016/j.jcms.2022.02.002}
}

@misc{zhou2018open3d,
  author    = {Zhou, Qian-Yi and Park, Jaesik and Koltun, Vladlen},
  title     = {{Open3D}: A modern library for {3D} data processing},
  year      = {2018},
  note      = {arXiv preprint arXiv:1801.09847},
  doi       = {10.48550/arXiv.1801.09847},
  eprint    = {1801.09847},
  archivePrefix = {arXiv}
}

@article{tangthaweesuk2025accuracy,
  author    = {Tangthaweesuk, Nichakun and Raocharernporn, Somchart},
  title     = {The accuracy of three-dimensional facial scan obtained from three different {3D} scanners},
  journal   = {PLOS ONE},
  volume    = {20},
  number    = {5},
  pages     = {e0322358},
  year      = {2025},
  doi       = {10.1371/journal.pone.0322358}
}

@article{young2016facial,
  author={Young, Nathan M. and Sherathiya, Krunal and Gutierrez, Luis and Nguyen, Emerald and Bekmezian, Sona and Huang, John C. and Hallgr{\'i}msson, Benedikt and Lee, Janice S. and Marcucio, Ralph S.},
  title={Facial surface morphology predicts variation in internal skeletal shape},
  journal={American Journal of Orthodontics and Dentofacial Orthopedics},
  volume={149},
  number={4},
  pages={501--508},
  year={2016},
  doi={10.1016/j.ajodo.2015.09.028}
}

@article{serafin2023accuracy,
  author={Serafin, Marco and Baldini, Benedetta and Cabitza, Federico and Carrafiello, Gianpaolo and Baselli, Giuseppe and Del Fabbro, Massimo and Sforza, Chiarella and Caprioglio, Alberto and Tartaglia, Gianluca M.},
  title={Accuracy of automated {3D} cephalometric landmarks by deep learning algorithms: systematic review and meta-analysis},
  journal={La radiologia medica}, volume={128}, number = {5}, 
  pages={544--555}, year={2023}, doi={10.1007/s11547-023-01629-2}
}

@misc{wang2025bfsm,
  author    = {Wang, Zidu and Xu, Meng and Xu, Miao and Ma, Hengyuan and Zhao, Jiankuo and Li, Xutao and Zhu, Xiangyu and Lei, Zhen},
  title     = {{BFSM}: {3D} bidirectional face--skull morphable model},
  year      = {2025},
  note      = {arXiv preprint arXiv:2509.24577},
  doi       = {10.48550/arXiv.2509.24577},
  eprint    = {2509.24577},
  archivePrefix = {arXiv}
}

@article{zhang2025tcfnet,
  author={Zhang, Runshi and Jie, Bimeng and He, Yang and Wang, Junchen},
  title={{TCFNet}: Bidirectional Face-Bone Transformation via a Transformer-based Coarse-to-Fine Point Movement Network},
  journal={Medical Image Analysis}, volume={105}, pages={103653}, year={2025}, doi={10.1016/j.media.2025.103653}
}

@article{heike2010stereophotogrammetry,
  author={Heike, Carrie L. and Upson, Kristen and Stuhaug, Erik and Weinberg, Seth M.},
  title={{3D} digital stereophotogrammetry: a practical guide to facial image acquisition},
  journal={Head \& Face Medicine}, volume={6}, pages={18}, year={2010}, doi={10.1186/1746-160X-6-18}
}

@article{ozsoy2015position,
  author={Ozsoy, Umut and Sekerci, Rahime and Ogut, Eren},
  title={Effect of sitting, standing, and supine body positions on facial soft tissue: detailed {3D} analysis},
  journal={International Journal of Oral and Maxillofacial Surgery}, volume={44}, number={10}, pages={1309--1316}, year={2015}, doi={10.1016/j.ijom.2015.06.005}
}

@misc{canfield2026vectrah1,
  author       = {{Canfield Scientific}},
  title        = {{VECTRA H1 3D Imaging System}},
  year         = {2026},
  howpublished = {Product webpage, \url{https://air.canfieldsci.com/imaging-systems/vectra-h1-3d-imaging-system/}},
  note         = {Accessed 18 July 2026}
}

@article{schlicher2012consistency,
  title={Consistency and precision of landmark identification in three-dimensional cone beam computed tomography scans},
  author={Schlicher, Will and Nielsen, Ib and Huang, John C. and Maki, Koutaro and Hatcher, David C. and Miller, A. J.},
  journal={European Journal of Orthodontics},
  volume={34},
  number={3},
  pages={263--275},
  year={2012},
  doi={10.1093/ejo/cjq144}
}

@article{hassan2013precision,
  title={Precision of identifying cephalometric landmarks with cone beam computed tomography in vivo},
  author={Hassan, Bassam and Nijkamp, Peter and Verheij, Hans and Tairie, Jamshed and Vink, Christian and van der Stelt, Paul and van Beek, Herman},
  journal={European Journal of Orthodontics},
  volume={35},
  number={1},
  pages={38--44},
  year={2013},
  doi={10.1093/ejo/cjr050}
}

@article{sam2019reliability,
  title={Reliability of different three-dimensional cephalometric landmarks in cone-beam computed tomography: A systematic review},
  author={Sam, Alycia and Currie, Kris and Oh, Heesoo and Flores-Mir, Carlos and Lagrav{\`e}re-Vich, Manuel},
  journal={The Angle Orthodontist},
  volume={89},
  number={2},
  pages={317--332},
  year={2019},
  doi={10.2319/042018-302.1}
}

@article{kim2022reliability,
  title={Reliability of cephalometric landmark identification on three-dimensional computed tomographic images},
  author={Kim, Jung-Hoon and An, SangIn and Hwang, Dong-Min},
  journal={British Journal of Oral and Maxillofacial Surgery},
  volume={60},
  number={3},
  pages={320--325},
  year={2022},
  doi={10.1016/j.bjoms.2021.07.003}
}

\clearpage
\appendix
\setcounter{section}{0}
\setcounter{subsection}{0}
\setcounter{figure}{0}
\setcounter{table}{0}
\setcounter{equation}{0}
\renewcommand{\thesection}{S\arabic{section}}
\renewcommand{\thesubsection}{S\arabic{section}.\arabic{subsection}}
\renewcommand{\thefigure}{S\arabic{figure}}
\renewcommand{\thetable}{S\arabic{table}}
\renewcommand{\theequation}{S\arabic{equation}}
\renewcommand{\theHsection}{supp.\arabic{section}}
\renewcommand{\theHsubsection}{supp.\arabic{section}.\arabic{subsection}}
\renewcommand{\theHfigure}{supp.\arabic{figure}}
\renewcommand{\theHtable}{supp.\arabic{table}}
\renewcommand{\theHequation}{supp.\arabic{equation}}
\markboth{Supplementary Material}{Supplementary Material}
\pdfbookmark[0]{Supplementary Material}{supplement}
\begin{center}
{\LARGE\bfseries Supplementary Material}\par
\vspace{0.5em}
{\large Surface-to-Skeleton 3D Cephalometry}\par
\end{center}
\vspace{1em}
\suppressfloats[t]
\section{Purpose of the Supplement}
This supplement expands the evidence and implementation details for the main submission. Its main role is to make the landmark definitions and anatomical grouping, patient-specific-signal controls, coverage dependence, optical-transfer boundary, observer-reliability limitation, and failure analysis easier to audit. It then reports the fixed implementation details, validation-selected ablations, final automatic ShapeGate variant, optical-like stress operators, preoperative Vectra transfer pilot, and coverage-standardized CT-only retraining diagnostic for the integrated hierarchical point-cloud model.

The model definitions, evaluation scripts, synthetic degradation operators, patient-mismatch control script, and table-generation code are publicly available.\footnote{\url{https://github.com/tomoki-pc/surface-to-skeleton-3d-cephalometry}} The retrospective clinical CT data are not publicly released in this study, but the fixed evaluation protocol and implementation details are provided to support independent reimplementation.
\section{Landmark Set and Anatomical Observability}
Table~\ref{tab:supp_landmarks} lists all 24 targets. The grouping is based on how directly an external soft-tissue surface can be expected to constrain each landmark, motivated by prior evidence that external facial morphology covaries with internal skeletal shape while recognizing that landmark-level surface observability varies \cite{young2016facial}. G0 points are on the visible surface, G1 skeletal points have strong frontal soft-tissue correlates, G2 points are lateral and more affected by soft-tissue thickness and field-of-view variation, and G3 points are deep or surface-invisible. Table~\ref{tab:supp_landmark_definitions} provides the annotation definitions used for each landmark.

\begin{table}[t]
\caption{Landmarks, target modality, and observability group. L/R denotes left/right.}
\label{tab:supp_landmarks}
\centering
\scriptsize
\setlength{\tabcolsep}{3.0pt}
\begin{tabularx}{\linewidth}{lXlX}
\toprule
Group & Landmarks & Target & Rationale\\
\midrule
G0 Surface & Prn, Sn, Pog$'$ & Soft tissue & Directly defined on the input surface\\
G1 Frontal & Na, Or L/R, Pt.A, ANS, Pt.B, Pog, Me & Skeletal & Frontal skeletal points with clearer surface correlates\\
G2 Lateral & AZ, ZA, Go L/R & Skeletal & Lateral sites with variable soft-tissue thickness and field-of-view variation\\
G3 Deep & S, Ba, FO L/R, PNS, Po L/R, CoS L/R & Skeletal & Deep or largely surface-invisible anatomy\\
\bottomrule
\end{tabularx}
\end{table}

\begin{table}[t]
\caption{Annotation definitions for the 24 target landmarks.}
\label{tab:supp_landmark_definitions}
\centering
\scriptsize
\setlength{\tabcolsep}{2.4pt}
\renewcommand{\arraystretch}{0.95}
\begin{tabularx}{\linewidth}{@{}p{0.30\linewidth}p{0.12\linewidth}X@{}}
\toprule
Landmark & Symbol & Definition used in annotation\\
\midrule
Anterior Nasal Spine & ANS & Tip of the anterior nasal spine\\
Right Zygomatic Arch Origin & AZ & Center point of the right zygomatic arch origin\\
Basion & Ba & Most inferior point on the anterior margin of the foramen magnum\\
Left Condylion Superius & CoS L & Superior point of the left mandibular condyle\\
Right Condylion Superius & CoS R & Superior point of the right mandibular condyle\\
Left Foramen Ovale & FO L & Left foramen ovale\\
Right Foramen Ovale & FO R & Right foramen ovale\\
Left Gonion & Go L & Point where the bisector of the angle formed by the mandibular lower-border plane and posterior ramus-border plane intersects the left gonial contour\\
Right Gonion & Go R & Point where the bisector of the angle formed by the mandibular lower-border plane and posterior ramus-border plane intersects the right gonial contour\\
Menton & Me & Lowest point on the mandibular symphyseal contour in the midsagittal section\\
Nasion & Na & Most anterior point of the frontonasal suture\\
Left Orbitale & Or L & Lowest point on the left orbital rim\\
Right Orbitale & Or R & Lowest point on the right orbital rim\\
Posterior Nasal Spine & PNS & Tip of the posterior nasal spine\\
Left Porion & Po L & Uppermost point of the left external auditory meatus\\
Right Porion & Po R & Uppermost point of the right external auditory meatus\\
Pogonion & Pog & Most anterior point on the mandibular symphyseal contour in the midsagittal section\\
A-Point & Pt.A & Deepest point on the maxillary contour between ANS and the alveolar crest between the maxillary central incisors\\
Sella & S & Center of the sella turcica\\
Left Zygomatic Arch Origin & ZA & Center point of the left zygomatic arch origin\\
B-Point & Pt.B & Deepest point on the mandibular contour between the alveolar crest between the mandibular central incisors and Pogonion\\
Pronasale & Prn & Tip of the nose\\
Subnasale & Sn & Subnasal point below the nose\\
Soft Pogonion & Pog$'$ & Most protrusive point on the soft-tissue chin surface\\
\bottomrule
\end{tabularx}
\end{table}

\section{Per-Landmark Results}
Table~\ref{tab:supp_perlm} gives the per-landmark mean radial error and the mean learned ShapeGate value on the 40 held-out cases for the reported no-point-context ShapeGate variant. The most visible soft-tissue points, Prn and Sn, have low error. Larger gates for PNS, S, and CoS indicate that ShapeGate often relies more strongly on the PCA-consistent anatomical correction for surface-invisible landmarks, although gates are learned without landmark-specific ranges and should be interpreted as correction strengths rather than direct observability labels.

\begin{table}[!htbp]
\caption{Per-landmark MRE and mean learned gate (40 test cases; ordered by G0--G3).}
\label{tab:supp_perlm}
\centering
\small
\setlength{\tabcolsep}{4.0pt}
\begin{tabular}{lcclcc}
\toprule
Landmark & MRE & Gate & Landmark & MRE & Gate\\
 & (mm) & & & (mm) & \\
\midrule
Prn & 1.53 & 0.08 & ZA & 3.39 & 0.33\\
Sn & 1.52 & 0.25 & Go L & 3.81 & 0.45\\
Pog$'$ & 4.63 & 0.19 & Go R & 3.37 & 0.29\\
Na & 3.12 & 0.08 & S & 3.05 & 0.57\\
Or L & 2.29 & 0.38 & Ba & 3.11 & 0.46\\
Or R & 2.72 & 0.44 & FO L & 3.07 & 0.33\\
Pt.A & 3.10 & 0.48 & FO R & 2.97 & 0.53\\
ANS & 2.00 & 0.38 & PNS & 2.17 & 0.59\\
Pt.B & 2.98 & 0.41 & Po L & 2.97 & 0.39\\
Pog & 2.21 & 0.48 & Po R & 3.19 & 0.39\\
Me & 2.61 & 0.39 & CoS L & 3.36 & 0.60\\
AZ & 3.42 & 0.36 & CoS R & 3.37 & 0.56\\
\bottomrule
\end{tabular}
\end{table}

\section{Qualitative Locked-Test Examples}
\begin{figure}[t]
\centering
\includegraphics[width=\linewidth]{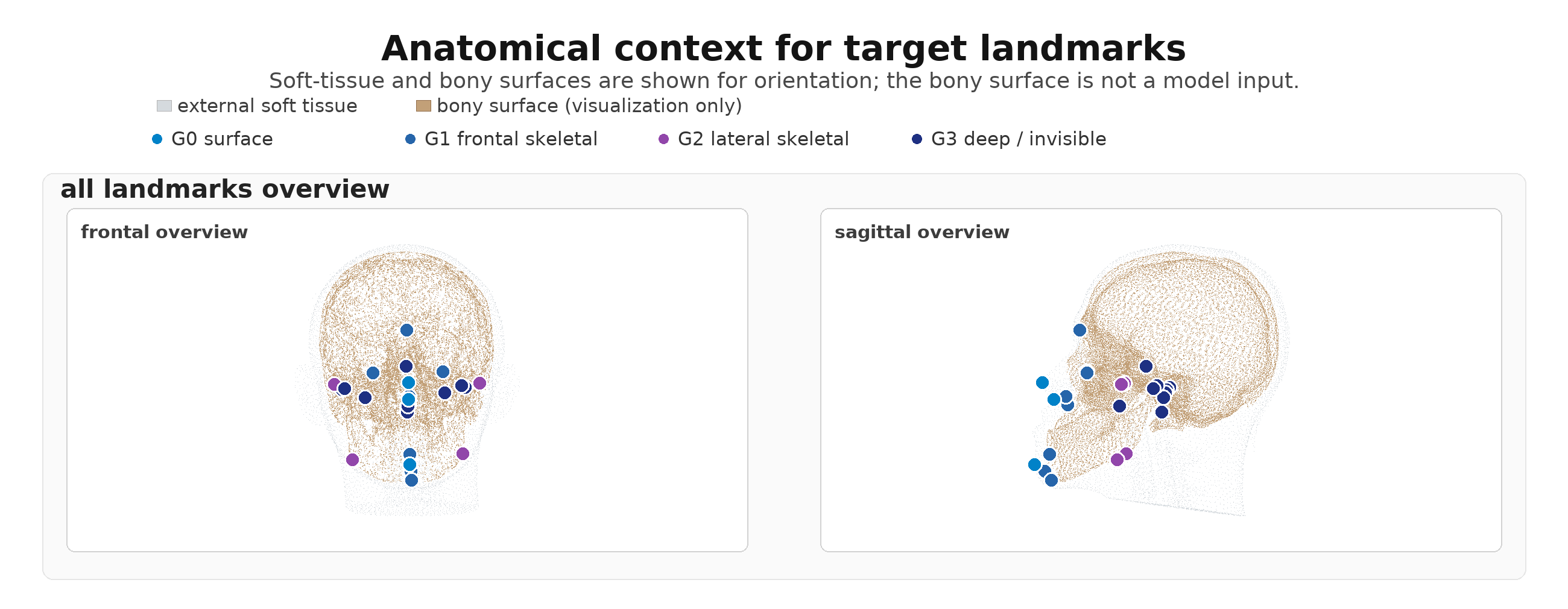}
\caption{Anatomical overview of all target landmarks on one representative
locked-test case. External soft tissue and bony surface projections are shown
only for orientation; the bony surface is not a model input. No patient
identifier is shown.}
\label{fig:supp_landmark_overview}
\end{figure}

\begin{figure}[tp]
\centering
\includegraphics[width=\linewidth]{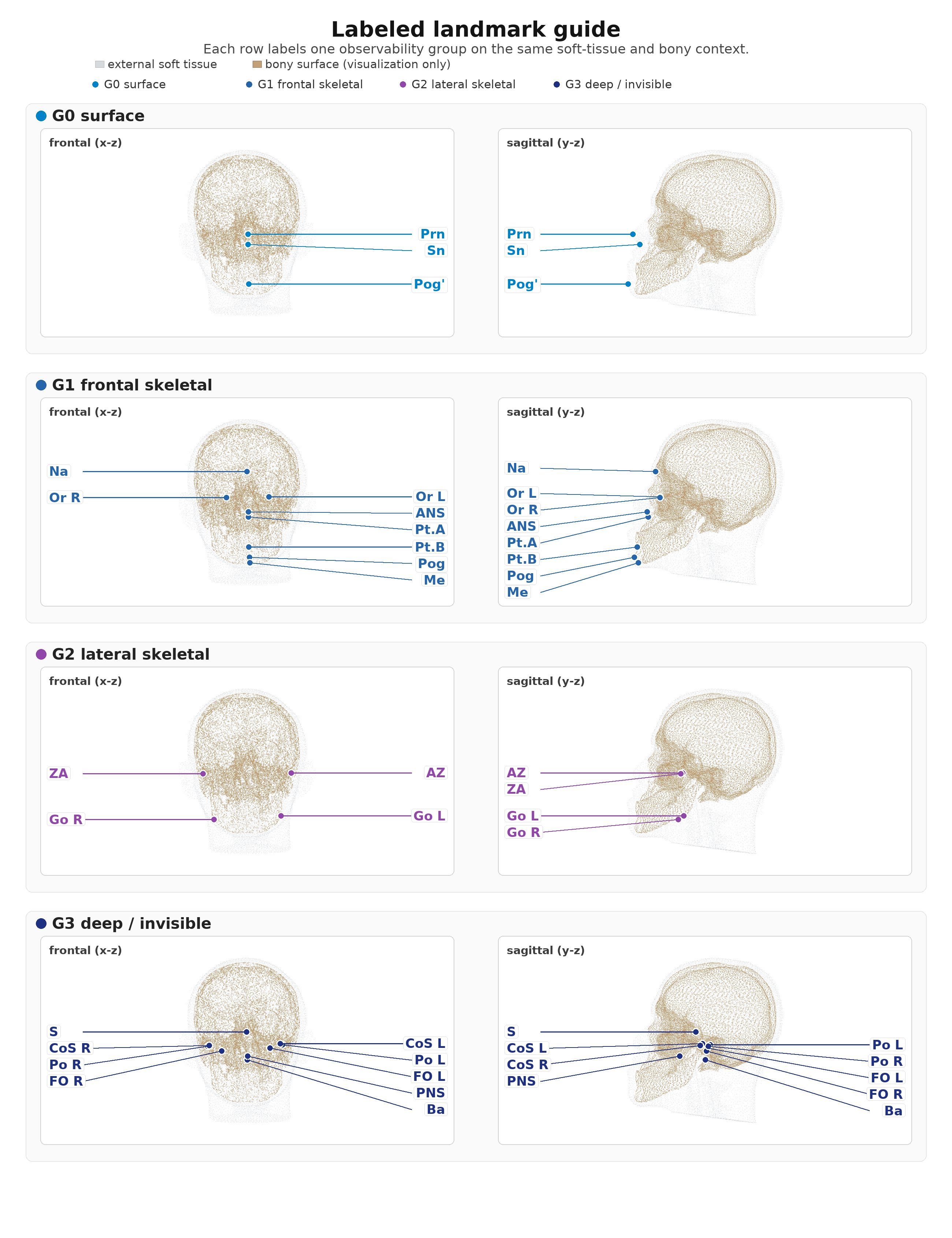}
\caption{Landmark-location guide for the qualitative overlays. The guide shows
reference landmark positions on one representative locked-test case,
separated by observability group and shown in frontal and sagittal projections.
External soft tissue and bony surface projections are shown only for
orientation; the bony surface is not a model input. No patient identifier is
shown.}
\label{fig:supp_landmark_guide}
\end{figure}

\begin{figure}[tp]
\centering
\includegraphics[width=\linewidth]{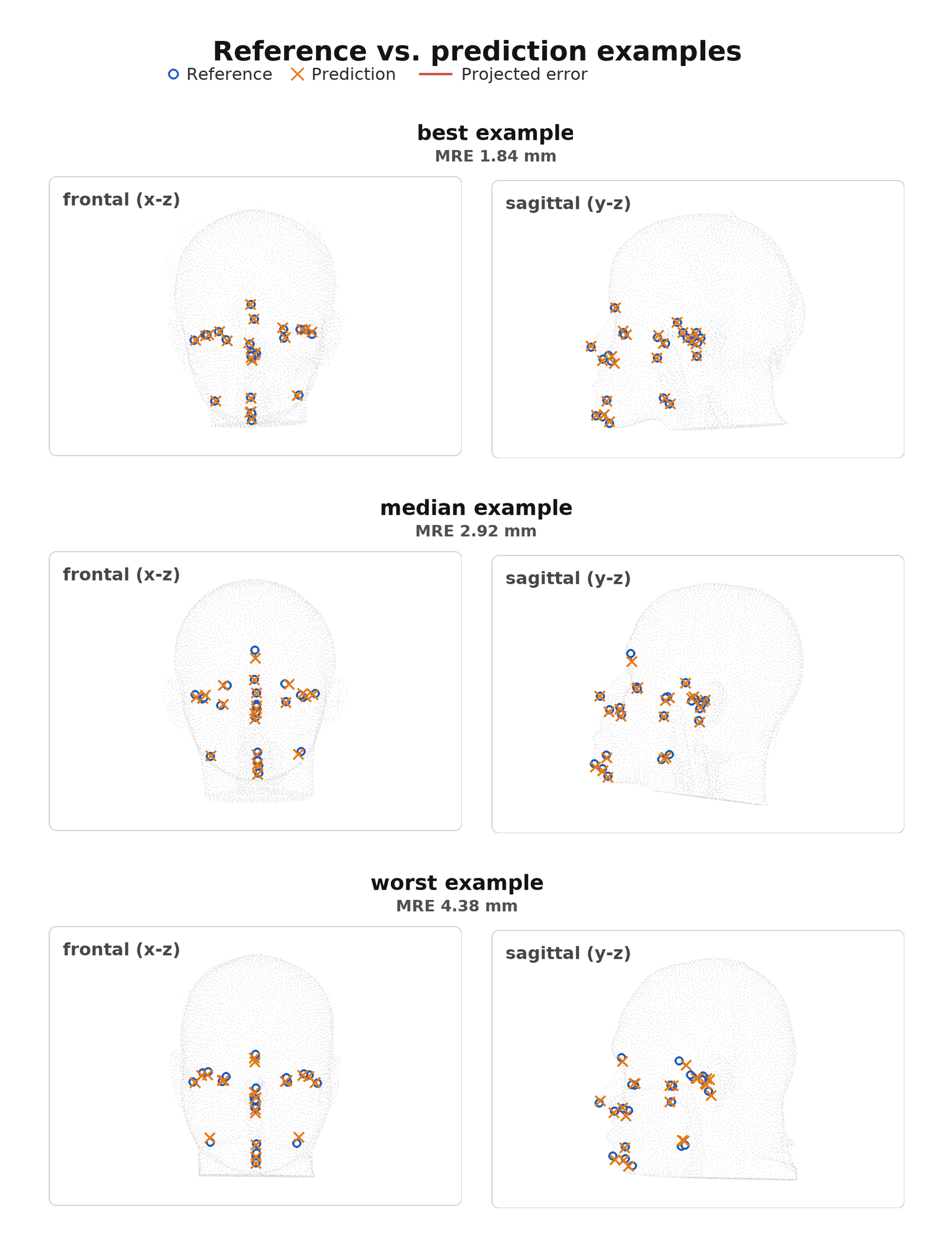}
\caption{Representative qualitative projections for the locked CT-derived test
set. Blue circles denote reference landmark positions and orange crosses
denote predictions. Grey points show the external CT-derived soft-tissue
surface.}
\label{fig:supp_qual_examples}
\end{figure}

Figure~\ref{fig:supp_landmark_overview} first shows the target landmarks in
anatomical context, and Fig.~\ref{fig:supp_landmark_guide} then labels the same
landmark abbreviations by observability group. Figure~\ref{fig:supp_qual_examples}
shows representative best, median, and worst locked-test cases according to
patient-mean MRE. The panels are orthographic projections of the CT-derived
external surface with projected reference and predicted landmark positions.
Hidden skeletal landmarks are therefore shown in projection rather than as
points lying on the visible surface. The examples are intended to illustrate the
spatial pattern of errors; quantitative interpretation should rely on the
tables in the main paper.

\section{Patient-Mismatch Control}
We used a frozen post-hoc patient-mismatch control to test whether the final
model relies on matched patient-specific surface geometry rather than only a
population or landmark-shape prior. The full model was not retrained or
reselected. Instead, for each locked-test case, we used the saved final
prediction produced from another patient's CT-derived external surface and
computed errors against the original patient's finalized reference landmarks.
We used 10 deterministic non-identity permutations, each constrained to have no
self-matches, and averaged the resulting metrics. Thus this control changes
only the pairing between surface-derived predictions and reference landmarks;
it does not change the trained model or the locked test set.

As an analysis-only control for global pose and scale confounding, we also
aligned each mismatched 24-landmark prediction configuration to the target
reference configuration by least-squares similarity transformation
(translation, rotation, and uniform scale; no reflection) before recomputing
errors. This alignment uses the target reference landmarks, so it is not a
deployable evaluation metric. Its purpose is only to test whether the raw
mismatch result is explained by global similarity differences between patients.

The raw mismatch control increased overall MRE from 2.91 to 11.87~mm
(SD 0.55), which is worse than the training-mean landmark baseline reported in
the main paper. Group-wise MRE increased for every observability group:
G0 from 2.56 to 12.65~mm, G1 from 2.63 to 12.22~mm, G2 from 3.50 to
11.99~mm, and G3 from 3.03 to 11.25~mm
(Table~\ref{tab:supp_mismatch}). After reference-based similarity alignment,
the mismatched MRE decreased to 5.36~mm overall, but remained above the clean
matched result for all groups (G2 5.54~mm; G3 4.35~mm). The G2 result is
important because lateral landmarks were the hardest group in the clean
evaluation and have weak external-surface signatures. These degradations
indicate that the reported clean-test performance is not explained by a fixed
population landmark configuration, the PCA shape prior alone, or global
similarity differences alone. They also do not imply that the model recovers an
observer-independent anatomical truth; the target remains the finalized
consensus coordinate system discussed in the main paper.

\begin{table}[t]
\caption{Patient-mismatch control on the locked test set. Matched is the original
patient pairing. Mismatched rows average
10 deterministic non-identity permutations of saved final predictions across
patients while keeping each patient's reference landmarks fixed. Similarity
alignment uses the target reference landmarks and is therefore an analysis-only
control, not a deployable metric.}
\label{tab:supp_mismatch}
\centering
\scriptsize
\setlength{\tabcolsep}{1.9pt}
\begin{tabular}{lccccccc}
\toprule
Condition & Pairings & MRE & G0 & G1 & G2 & G3 & Fail@10\\
 & & (mm) & (mm) & (mm) & (mm) & (mm) & (\%)\\
\midrule
Matched & -- & 2.91 & 2.56 & 2.63 & 3.50 & 3.03 & 0.7\\
Mismatched & 10 & 11.87$\pm$0.55 & 12.65$\pm$0.41 & 12.22$\pm$0.55 & 11.99$\pm$0.67 & 11.25$\pm$0.71 & 55.1\\
Mismatch+sim. & 10 & 5.36$\pm$0.16 & 6.36$\pm$0.15 & 6.03$\pm$0.24 & 5.54$\pm$0.21 & 4.35$\pm$0.14 & 6.7\\
\bottomrule
\end{tabular}
\end{table}

\section{Field-of-View Truncation Subgroup}
Table~\ref{tab:supp_fov} gives the exploratory locked-test stratification by
naturally occurring field-of-view truncation. These subgroups are small and were
defined after the final locked evaluation, so they are descriptive rather than
powered subgroup tests.

\begin{table}[t]
\caption{Exploratory performance by field-of-view truncation on the locked test set.}
\label{tab:supp_fov}
\centering
\small
\setlength{\tabcolsep}{3.2pt}
\begin{tabular}{lcccccc}
\toprule
Test subgroup & Cases & MRE & SDR@3 & SDR@5 & P95 & Fail@10\\
 & & (mm) & (\%) & (\%) & (mm) & (\%)\\
\midrule
Complete FOV & 25 & 2.83 & 65.0 & 88.3 & 6.33 & 0.5\\
Superior truncation & 6 & 3.20 & 56.3 & 88.9 & 6.43 & 1.4\\
Inferior truncation & 4 & 2.86 & 64.6 & 90.6 & 6.17 & 1.0\\
Both-end truncation & 5 & 3.03 & 57.5 & 83.3 & 6.73 & 0.8\\
\bottomrule
\end{tabular}
\end{table}

\section{Implementation Details}
Table~\ref{tab:supp_impl} summarizes the main fixed implementation choices. Stage-1, Stage-2, and Stage-2b are trained as separate modules. ShapeGate is trained only after generating out-of-fold Stage-2b predictions on the inner-training split, so its training input matches the error distribution expected at test time more closely than teacher-forced reference-coordinate inputs would.

\begin{table}[tp]
\caption{Fixed implementation and selection details.}
\label{tab:supp_impl}
\centering
\small
\setlength{\tabcolsep}{3.5pt}
\begin{tabularx}{\linewidth}{lX}
\toprule
Component & Setting\\
\midrule
Input & 8,192 registered external-surface points with outward normals; coordinates divided by 100~mm for network input and converted back to millimeters for evaluation; coordinates and normals only\\
Registration & Geometry renderings from 39 views, MediaPipe facial keypoints \cite{lugaresi2019mediapipe,kartynnik2019facial}, rigid initialization, then point-to-plane ICP in Open3D \cite{zhou2018open3d}; no target cephalometric landmarks used\\
Stage 1 & Sonata/PTv3 encoder \cite{wu2025sonata,wu2024ptv3} initialized from public \texttt{facebook/sonata} weights and fully fine-tuned; no local study-specific Sonata checkpoint or cohort pretraining was used; per-point offsets and confidences; 150 epochs; AdamW; cosine decay; encoder LR $10^{-5}$, head LR $10^{-3}$\\
Stage 2 & Landmark-specific 50-mm PointNeXt ROI \cite{qian2022pointnext} using relative coordinates, normals, and landmark identity; at most 4,096 points; fewer than 128 ROI points retain Stage-1 estimate; residual damping $\alpha=0.75$ except Sella ($\alpha=0$)\\
Stage 2b & 80-mm PointNeXt ROI \cite{qian2022pointnext} for 14 high-error landmarks using the same ROI feature convention; at most 8,192 points; residual damping 0.4; all other landmarks retain Stage-2 estimates\\
ShapeGate & Reported no-point-context variant; PCA shape model \cite{cootes1995active,tothova2018probabilistic} fitted on 160 inner-training landmark configurations; 16 components explain approximately 92.9\% of the inner-training variance; 24 landmark tokens; two self-attention layers; one shared scalar-output head yielding one sigmoid gate per landmark token in $[0,1]$; no free XYZ residual\\
Split hygiene & Architecture and hyperparameters were selected on the 160/40 inner split. Final Stage-1, Stage-2, and Stage-2b modules were retrained on all 200 training cases after choices were fixed. PCA and ShapeGate were fitted on the 160 inner-training cases with 40 inner-validation cases for early stopping; the 40 held-out test cases were not used for fitting or checkpoint selection\\
Augmentation & Small rigid perturbations, Gaussian coordinate noise, point dropout, and ROI-center jitter\\
Model selection & Architecture and hyperparameters selected on the 160/40 inner split; the final no-point-context ShapeGate used a uniform initial gate value of 0.10 and not-worse/stay-close weights selected from validation and out-of-fold behavior; the same gate parameterization was used for all landmarks\\
\bottomrule
\end{tabularx}
\end{table}

\paragraph{Stage-2 Residual-Damping Selection.}
For the 50-mm refiner, candidate values $\alpha=0$, $0.25$, $0.50$,
$0.75$, and $1.00$ yielded inner-validation all-landmark MREs of 3.766,
3.481, 3.292, 3.217, and 3.271~mm, respectively, fixing $\alpha=0.75$ for 23 landmarks. For
Sella, the full residual increased MRE from 4.419 to 4.787~mm, so its
residual was suppressed ($\alpha_{\mathrm S}=0$).
For Stage~2b, 80- and 90-mm ROIs and $\alpha=0,0.025,\ldots,1.00$ were compared on inner validation. The 80-mm, $\alpha=0.40$ setting minimized MRE (3.1419~mm); neighboring values 0.375 and 0.425 yielded 3.1421 and 3.1423~mm, respectively, indicating low local sensitivity.

\paragraph{Global-Descriptor PCA-Ridge Shape-Model Baseline.}
The statistical-shape baseline in the main comparison table was fitted only on the 200 training cases. We fit PCA to flattened 72-dimensional landmark configurations, following standard statistical shape-model practice \cite{cootes1995active,tothova2018probabilistic}, and retained 16 components, explaining 95.2\% of the training variance. For each aligned external-surface point cloud, we computed a fixed 57-dimensional global descriptor from the same 8,192 input points: coordinate centroid, standard deviation, minima, maxima, extents, per-axis quantiles, covariance entries, and radial quantiles around the centroid. Ridge regression predicted the 16 PCA scores from these descriptors; the regularization parameter was selected from $\{0,0.01,0.1,1,10,100,1000\}$ by deterministic five-fold cross-validation on the 200 training cases. No held-out test case was used for PCA fitting, descriptor normalization, regression fitting, or regularization selection. This baseline is intended as a simple case-specific statistical shape model, not as a neural point-cloud model or as a standalone mean-shape predictor.

\paragraph{Registration Details.}
The alignment script rendered 39 geometry views per case, using 13 azimuths ($-30^{\circ}$ to $30^{\circ}$ in $5^{\circ}$ steps) and three elevations ($-20^{\circ}$, $0^{\circ}$, $20^{\circ}$). MediaPipe detections \cite{lugaresi2019mediapipe,kartynnik2019facial} were associated with the mesh by multi-view consistency filtering and summarized as 3D mean landmark points. For initialization, MediaPipe IDs were retained only when they had at least 30 supporting views, root-mean-square distance to their 3D mean of at most 8~mm, maximum distance to their mean of at most 25~mm, and at least 30 common IDs with the template. A rigid initial transform was estimated from common 3D MediaPipe landmarks by SVD after trimming the worst 20\% landmark correspondences. The final alignment sampled 30,000 surface points, voxel-downsampled with 2-mm voxels, estimated normals, and ran Open3D point-to-plane ICP \cite{zhou2018open3d} with a 10-mm correspondence threshold and 80 iterations. The resulting rigid transform was applied identically to the soft-tissue mesh, bony mesh, and landmark CSV.

\paragraph{PCA Projection and ShapeGate Fitting.}
The reported ShapeGate run used the center PCA mode. We divided the 160 inner-training reference landmark configurations by 100~mm, centered each by its own centroid, flattened them to 72 dimensions, and used them to fit a PCA shape model \cite{cootes1995active,tothova2018probabilistic}. At inference, each Stage-2b prediction was transformed in the same way using its own centroid, projected onto the 16-component PCA subspace, and inverse-centered with that same centroid before computing the prior correction $\Delta_k$. No per-case RMS-radius normalization was used in the reported final run. Thus the final point-cloud predictors were retrained on all 200 training cases after model choices were fixed, whereas PCA and ShapeGate retained the inner-training/inner-validation protocol shown in Table~\ref{tab:supp_impl}; the 40 held-out test cases were not used for PCA fitting, ShapeGate fitting, or early stopping.

\paragraph{Automatic Gate Parameterization.}
The reported model applies one shared head to every landmark token; its scalar output is a case- and landmark-specific gate. Thus, the head parameters and mapping are shared across landmarks, but a single common gate value is not imposed. For landmark $k$, the raw gate logit $a_k$ is mapped only by a sigmoid,
\begin{equation}
g_k=\sigma(a_k),
\end{equation}
so the same $[0,1]$ range applies to every landmark. The reported model does not assign landmark-specific gate limits or initialization values, and no auxiliary gate regularizer is used. The only initialization prior is a uniform gate bias corresponding to an initial value of 0.10. The final position is therefore still constrained to lie on the line segment between the Stage-2b prediction and its PCA projection, but the amount of correction is learned automatically from the landmark-token features.

Table~\ref{tab:supp_gate_sweep} summarizes the automatic-gate sweep. The reported setting was selected using inner-validation and out-of-fold performance; the held-out test set was not used to compare gate candidates. All rows use the same shared $[0,1]$ sigmoid gate for every landmark. Table~\ref{tab:supp_loss_weights} lists the loss terms and weights used by the final reported modules.

For the fixed-scalar ablation, the PCA mode and 16-component subspace were held fixed and a single gate shared across all cases and landmarks was selected from $g\in\{0,0.05,0.10,0.20,0.30,0.50,0.75,1.00\}$. Inner validation selected $g=0.50$ (3.036~mm MRE), which was frozen before locked-test evaluation. It achieved 2.937~mm MRE, 61.9\% SDR@3, and 87.7\% SDR@5 on the test set. The patient-level fixed-minus-learned difference was 0.023~mm (95\% CI 0.011--0.035; $p<.001$); although statistically detectable, this small absolute effect does not establish clinical importance.

\begin{table}[t]
\caption{Automatic-gate sweep. The reported setting uses a shared $[0,1]$ sigmoid gate for every landmark. Bold indicates the best inner-validation and out-of-fold values used for selection.}
\label{tab:supp_gate_sweep}
\centering
\scriptsize
\setlength{\tabcolsep}{3.4pt}
\begin{tabular}{lccccc}
\toprule
Config. & Init. & Not-worse & Stay & Val & OOF\\
 & $g$ & weight & weight & MRE & MRE\\
\midrule
Default initialization & 0.25 & 10 & 5 & 3.018 & 3.536\\
Lower-gate initialization (reported) & 0.10 & 10 & 5 & \textbf{3.014} & \textbf{3.524}\\
Higher not-worse penalty & 0.25 & 20 & 5 & 3.021 & 3.542\\
Higher stay-close penalty & 0.25 & 10 & 10 & 3.040 & 3.577\\
Higher not-worse and stay-close penalties & 0.25 & 20 & 10 & 3.042 & 3.582\\
\bottomrule
\end{tabular}
\end{table}

\begin{table}[t]
\caption{Loss weights used by the final reported modules.}
\label{tab:supp_loss_weights}
\centering
\small
\setlength{\tabcolsep}{3.5pt}
\begin{tabularx}{\linewidth}{lX}
\toprule
Module & Loss terms and weights\\
\midrule
Stage 1 and ROI refiners & Coordinate prediction, proposal supervision, distance-based confidence, and entropy weights were 1.0, 0.3, 0.02, and 0.0, respectively\\
ShapeGate coordinate term & Smooth-$\ell_1$ final-position loss with 1.0-mm beta\\
ShapeGate over-threshold term & Squared error above the 3-mm target threshold; weight 1.0; squared errors were computed in millimeters and divided by $100^2$ to match the coordinate normalization used by the network\\
ShapeGate not-worse term & Squared excess over the Stage-2b input error, margin 0~mm; weight 10.0; squared errors were computed in millimeters and divided by $100^2$\\
ShapeGate stay-close term & Squared correction magnitude; weight 5.0; squared correction magnitudes were computed in millimeters and divided by $100^2$\\
Unused pair-distance term & Weight 0.0 in the reported run\\
\bottomrule
\end{tabularx}
\end{table}

\section{Optical-Like Stress Operators and Paired Effects}
Table~\ref{tab:supp_degrade_ops} details the frozen-model degradation operators,
and Table~\ref{tab:supp_degradation_results} reports the corresponding stage-wise
errors and sparse-input sweep. These operators are intentionally described as
stress tests rather than scanner simulation. They change only the point cloud
and normals at inference; reference landmarks and model weights remain
unchanged. The face-only and tight face-only crops are all-40 analysis-only
coverage ablations that use synthetic percentile masks to remove posterior,
lateral, and inferior/superior coverage; they are not deployable optical-scan
preprocessing.

\begin{table}[t]
\caption{Synthetic input degradations used for the frozen-model sensitivity analysis.}
\label{tab:supp_degrade_ops}
\centering
\small
\setlength{\tabcolsep}{3.5pt}
\begin{tabularx}{\linewidth}{lX}
\toprule
Condition & Operator\\
\midrule
Clean & Registered CT-derived external surface, sampled to 8,192 points\\
Frontal crop & Keep points with anterior score above 30th percentile and absolute lateral score below 96th percentile; resample to 8,192 points\\
Face-only crop & Keep points with anterior score above 45th percentile and absolute lateral score below 90th percentile; remove inferior 8\% in $z$; resample to 8,192 points\\
Tight face-only crop & Keep points with anterior score above 55th percentile and absolute lateral score below 84th percentile; remove inferior 12\% and superior 2\% in $z$; resample to 8,192 points\\
Holes + noise & Remove six random spherical holes of 12~mm radius while retaining at least 1,024 points; add Gaussian coordinate noise ($\sigma=0.75$~mm) and normal noise ($\sigma=0.05$)\\
Combined & Stricter crop using 34th anterior percentile and 93rd lateral percentile; remove inferior 8\% in $z$; remove eight 10-mm holes with a 512-point minimum; add coordinate noise ($\sigma=0.75$~mm) and normal noise ($\sigma=0.06$); apply 2,048-point sparsification\\
Sparse 4,096/2,048/1,024 & Randomly select the specified number of unique points and resample with replacement to 8,192 points\\
\bottomrule
\end{tabularx}
\end{table}

\begin{table}[t]
\caption{Frozen-model optical-like sensitivity and sparse-input sweep on the
40 locked-test cases. Errors are in millimeters, and $\Delta$ is relative to
clean input.}
\label{tab:supp_degradation_results}
\centering
\scriptsize
\setlength{\tabcolsep}{3.0pt}
\begin{tabular}{lccccccc}
\toprule
Input & S1 & S2 & S2b & Final & $\Delta$ & SDR@5 & P95\\
 & & & & & & (\%) & (mm)\\
\midrule
Clean & 3.61 & 3.09 & 3.01 & 2.91 & -- & 88.0 & 6.50\\
Frontal crop & 4.31 & 3.48 & 3.36 & 3.21 & +0.30 & 83.8 & 6.82\\
Holes + noise & 4.10 & 3.66 & 3.48 & 3.28 & +0.36 & 83.2 & 7.17\\
Combined & 7.02 & 5.31 & 5.04 & 4.60 & +1.69 & 67.4 & 10.15\\
Sparse 4,096 & 5.27 & 4.24 & 3.98 & 3.69 & +0.78 & 78.5 & 7.75\\
Sparse 2,048 & 12.57 & 8.20 & 7.39 & 6.61 & +3.69 & 40.8 & 13.95\\
Sparse 1,024 & 32.45 & 21.09 & 18.88 & 17.64 & +14.73 & 2.2 & 33.87\\
\bottomrule
\end{tabular}
\end{table}

Table~\ref{tab:supp_paired_effects} reports patient-level paired effect sizes.
For model comparisons, the difference is baseline MRE minus full-pipeline MRE,
so positive values indicate improvement. For coverage ablations, the difference
is condition MRE minus clean MRE, so positive values indicate degradation. CIs
are 10,000-sample bootstrap intervals over the 40 patient-level paired
differences; $p$ values use 100,000 random sign flips and are descriptive for
the post-hoc coverage analyses. The bootstrap intervals and sign-flip
permutation tests follow standard resampling practice
\cite{efron1979bootstrap,good2005permutation}.

\begin{table}[t]
\caption{Patient-level paired effect sizes on the locked test set. Differences are in mm and computed after averaging errors over the 24 landmarks within each patient.}
\label{tab:supp_paired_effects}
\centering
\scriptsize
\setlength{\tabcolsep}{2.7pt}
\begin{tabular}{lccccc}
\toprule
Comparison & Direction & Mean diff. & 95\% CI & $d_z$ & $p$\\
\midrule
Stage 1 vs full & Baseline--full & 0.69 & 0.56--0.83 & 1.50 & $<10^{-4}$\\
50-mm ROI vs full & Baseline--full & 0.18 & 0.13--0.22 & 1.11 & $<10^{-4}$\\
Stage-2b vs full & Baseline--full & 0.10 & 0.06--0.13 & 0.92 & $<10^{-4}$\\
Fixed scalar gate vs full & Baseline--full & 0.023 & 0.011--0.035 & 0.60 & $<.001$\\
PTv3+ROI comparator vs full & Baseline--full & 0.22 & 0.14--0.31 & 0.77 & $<10^{-4}$\\
Frontal crop minus clean & Condition--clean & 0.30 & 0.19--0.42 & 0.80 & $<10^{-4}$\\
Face-only crop minus clean & Condition--clean & 0.82 & 0.51--1.20 & 0.73 & $<10^{-4}$\\
Tight face-only crop minus clean & Condition--clean & 1.86 & 1.37--2.45 & 1.06 & $<10^{-4}$\\
\bottomrule
\end{tabular}
\end{table}

\section{Uncertainty and Abstention Diagnostics}
We evaluated whether frozen-model internal signals could rank high-error cases
without training an additional test-set classifier. A case was labeled
high-error when its 24-landmark MRE exceeded 3~mm (17/40 cases). These results
are exploratory and intended only to motivate future manual-review or
abstention strategies. Table~\ref{tab:supp_uncertainty} reports the high-error
ranking signals and retention metrics.

\begin{table}[t]
\caption{Exploratory high-error ranking and abstention on the locked test set.}
\label{tab:supp_uncertainty}
\centering
\small
\begin{minipage}[t]{0.50\linewidth}
\centering
\setlength{\tabcolsep}{3.0pt}
\begin{tabular}{lcc}
\toprule
\multicolumn{3}{c}{High-error ranking}\\
\midrule
Signal & AUROC & 95\% CI\\
\midrule
Correction mean & 0.80 & 0.65--0.92\\
Shape delta mean & 0.80 & 0.65--0.92\\
Shape delta P95 & 0.79 & 0.64--0.91\\
Disagreement mean & 0.76 & 0.57--0.90\\
\bottomrule
\end{tabular}
\end{minipage}\hfill
\begin{minipage}[t]{0.46\linewidth}
\centering
\setlength{\tabcolsep}{3.0pt}
\begin{tabular}{lcc}
\toprule
\multicolumn{3}{c}{Mean-disagreement abstention}\\
\midrule
Coverage & Case MRE & High-error\\
\midrule
100\% & 2.91 & 42.5\%\\
80\% & 2.74 & 31.3\%\\
70\% & 2.69 & 28.6\%\\
50\% & 2.66 & 25.0\%\\
\bottomrule
\end{tabular}
\end{minipage}
\end{table}

\section{Preoperative Vectra Transfer Pilot}
A post-hoc pilot evaluated the same frozen full-pipeline model on 14 preoperative facial scans acquired with a VECTRA H1 handheld stereophotogrammetry system (Canfield Scientific) \cite{canfield2026vectrah1} from eight patients in the locked test set. These were all 14 scan directories in the locked-test Vectra-aligned folder that could be linked to held-out CT cases and exported successfully; the scan list was determined by availability and anonymized linkage, not by model error. Visual inspection indicated that these Vectra scans primarily captured the anterior facial surface, unlike the CT-derived soft-tissue surfaces used for training and clean testing. The model weights, ShapeGate checkpoint, and CT landmark targets were unchanged. CT and Vectra were acquired on different preoperative days. The CT protocol used maximum intercuspation, closed lips, upright head posture, and ear-rod stabilization; facial expression was not otherwise specified. Vectra imaging used maximum intercuspation, closed lips, and upright posture, with open-eye and closed-eye states acquired when available, but no ear rods or other rigid head fixation and no specified facial expression. Some patients lacked one of the eye-state captures, and acquisition order was unavailable, so duplicate scans should not be interpreted as a formal repeatability experiment. For inference, Vectra scans were processed with the same template-based registration pipeline as CT-derived surfaces; the paired CT surface was not used for Vectra-to-CT alignment before prediction. The Vectra scan was used only as the input point cloud, while the target landmarks remained the CT-derived coordinates from the paired preoperative CT. Vectra points and normals were sampled from the aligned STL meshes with the same mesh-sampling utility used for CT-derived surfaces, yielding 8,192 input points per scan. Paired CT surfaces were used only for target evaluation and post-hoc surface-proximity diagnostics. Therefore these values are apparent optical-to-CT errors: they include model error, residual registration error, scanner-domain shift, and non-rigid differences between acquisitions \cite{heike2010stereophotogrammetry,ozsoy2015position,tangthaweesuk2025accuracy}.

To separate front-only coverage from scanner-domain effects, we also created a diagnostic CT-derived Vectra-coverage input. For each aligned Vectra scan, paired CT-derived surface points were selected inside the 1st--99th percentile Vectra spatial envelope with 10-mm margins in $x/z$ and 15-mm margin in $y$; within this envelope, the 70\% of CT points nearest to the Vectra point cloud were retained and resampled to 8,192 points. This analysis uses paired Vectra and CT geometry to define an input mask, so it is not a deployable preprocessing step; it is only a coverage-gap decomposition.

\begin{table}[t]
\caption{Coverage/domain decomposition on the same eight locked-test patients with 14 Vectra scans. Patient-level summaries first average over landmarks and scans for each patient. Prediction shift is mean landmark displacement from the corresponding clean CT-derived prediction; scan-pair shift summarizes descriptive within-patient variability across available duplicate scans.}
\label{tab:supp_vectra_pilot}
\centering
\scriptsize
\setlength{\tabcolsep}{4.0pt}
\begin{tabular}{lccc}
\toprule
Input & Patient MRE & Pred. shift vs. clean & Scan-pair shift\\
 & (mm) & (mm) & (mm)\\
\midrule
Clean CT-derived & 2.79 & -- & --\\
CT-derived coverage crop & 3.79 & 2.13 & 1.14\\
Frozen full-pipeline Vectra & 10.96 & 10.27 & 3.67\\
\bottomrule
\end{tabular}
\end{table}

\begin{table}[t]
\caption{Stage progression for CT-derived Vectra-coverage crops and full-pipeline Vectra scan inputs. Values are landmark-instance MREs averaged over the 14 scan inputs; Table~\ref{tab:supp_vectra_pilot} uses patient-level aggregation and therefore gives slightly different summary values.}
\label{tab:supp_vectra_stages}
\centering
\scriptsize
\setlength{\tabcolsep}{3.8pt}
\begin{tabular}{lcccc}
\toprule
Stage & \multicolumn{2}{c}{MRE (mm)} & \multicolumn{2}{c}{SDR@5 (\%)}\\
\cmidrule(lr){2-3}\cmidrule(lr){4-5}
 & CT crop & Vectra & CT crop & Vectra\\
\midrule
Stage 1 & 4.41 & 12.40 & 65.8 & 7.1\\
Stage 2 ROI & 4.32 & 11.83 & 68.2 & 12.8\\
Stage 2b Ext14 & 4.18 & 11.36 & 72.3 & 13.7\\
ShapeGate & 3.83 & 10.86 & 74.4 & 13.1\\
\bottomrule
\end{tabular}
\end{table}

\begin{table}[t]
\caption{Diagnostic analysis of the Vectra transfer gap. Surface-proximity correlations use 14 scans and are therefore descriptive rather than confirmatory.}
\label{tab:supp_vectra_diagnostics}
\centering
\scriptsize
\setlength{\tabcolsep}{4.0pt}
\begin{tabular}{lc}
\toprule
Diagnostic quantity & Value\\
\midrule
Scan-level final MRE range & 6.61--19.08~mm\\
Stage-1 to final mean improvement & 1.53~mm\\
Stage-2 ROI applied fraction & 95.8\%\\
G0 surface-landmark MRE & 9.76~mm\\
G1 frontal-landmark MRE & 9.31~mm\\
G2 lateral-landmark MRE & 10.99~mm\\
G3 deep-landmark MRE & 12.55~mm\\
Corr. with Vectra-to-CT median NN distance & $r=0.67$\\
Corr. with CT-overlap-to-Vectra median NN distance & $r=0.77$\\
Corr. with CT-overlap-to-Vectra P90 NN distance & $r=0.70$\\
\bottomrule
\end{tabular}
\end{table}

We further performed a coverage-standardized retraining diagnostic to test
whether the large Vectra error was mainly caused by using face-only scans with a
model trained on broader CT-derived head coverage. For both CT-derived surfaces
and Vectra scans, we computed a MediaPipe face-mesh bounding box
\cite{lugaresi2019mediapipe,kartynnik2019facial} after template
alignment, retained triangles of the aligned soft-tissue STL whose centroids
fell inside the box, and sampled 8,192 points with area weighting. The CT
targets were unchanged. This avoids the duplicate-point distribution shift that
would occur if the bounding box were applied only after the existing 8,192-point
NPZ sampling. All 200 training CT cases, 40 locked-test CT cases, and 14
Vectra scans were generated successfully; no case required bounding-box
expansion. The median retained triangle counts were 72,727 for CT training,
69,912 for locked-test CT, and 54,636 for Vectra.
Only the 200 CT training cases were used to retrain Stage~1 and Stage~2.
The 40 locked-test CT cases and 14 Vectra scans were regenerated with the
same MediaPipe-bounding-box mesh-sampling procedure only for evaluation; no
Vectra scan was used for training, model selection, checkpoint selection, or
hyperparameter tuning.

\begin{table}[t]
\caption{Coverage-standardized MediaPipe-bounding-box retraining diagnostic. Errors are landmark-instance MREs in millimeters. The diagnostic retrains only Stage~1 and Stage~2 on the 200 CT training cases with mesh-sampled face-bounding-box inputs; CT MRE is evaluated on the 40 locked-test CT cases, Vectra metrics are evaluated on 14 scans from eight patients, and Stage~2b and ShapeGate were not rerun.}
\label{tab:supp_vectra_mpbbox_retrain}
\centering
\begingroup
\scriptsize
\setlength{\tabcolsep}{2.6pt}
\renewcommand{\arraystretch}{1.08}
\begin{tabular}{@{}>{\raggedright\arraybackslash}p{0.28\linewidth}>{\raggedright\arraybackslash}p{0.17\linewidth}cccc@{}}
\toprule
Condition & Stage & CT & Vectra & Vectra & Vectra\\
 & & MRE & MRE & SDR@5 & Fail@10\\
 & & (mm) & (mm) & (\%) & (\%)\\
\midrule
Original whole-surface & Stage~1 & 3.61 & 12.40 & 7.1 & 55.4\\
Original whole-surface & Stage~2 & 3.09 & 11.83 & 12.8 & 53.3\\
Original whole-surface & Final ShapeGate & 2.91 & 10.86 & 13.1 & 48.8\\
MediaPipe-bbox mesh-sampled retrain & Stage~1 & 5.01 & 7.94 & 28.9 & 28.3\\
MediaPipe-bbox mesh-sampled retrain & Stage~2 & 4.19 & 8.45 & 27.4 & 31.2\\
\bottomrule
\end{tabular}
\endgroup
\end{table}

\begin{table}[t]
\caption{Group-wise apparent Vectra-to-CT errors for the coverage-standardized retraining diagnostic. Values are scan-level landmark-instance MREs in millimeters over the 14 Vectra scans.}
\label{tab:supp_vectra_mpbbox_groups}
\centering
\scriptsize
\setlength{\tabcolsep}{4.0pt}
\begin{tabular}{lccccc}
\toprule
Condition & Overall & G0 surface & G1 frontal & G2 lateral & G3 deep\\
\midrule
Original final ShapeGate & 10.86 & 9.76 & 9.31 & 10.99 & 12.55\\
MediaPipe-bbox Stage~1 & 7.94 & 9.47 & 8.99 & 8.90 & 6.08\\
MediaPipe-bbox Stage~2 & 8.45 & 10.20 & 9.29 & 9.09 & 6.83\\
\bottomrule
\end{tabular}
\end{table}

The retraining diagnostic suggests two points. First, face-region
standardization substantially improved global Stage~1 Vectra transfer: Stage~1
Vectra MRE decreased from 12.40 to 7.94~mm, SDR@5 increased from 7.1\% to
28.9\%, Fail@10 decreased from 55.4\% to 28.3\%, and the same-stage
CT--Vectra summary gap decreased from 8.79 to 2.93~mm. Numerically,
approximately three-quarters of this observed summary-gap narrowing came from
lower Vectra error (4.46~mm) rather than higher CT error (1.40~mm). This
decomposition is descriptive rather than causal or paired because the CT and
Vectra summaries use different evaluation sets and aggregation units. At the
same time, retraining does not restore clean CT-derived performance:
locked-test CT Stage~1 MRE increased from 3.61 to 5.01~mm and Stage~2 MRE
increased from 3.09 to 4.19~mm. This is consistent with the all-case
anterior-coverage ablation: removing posterior, lateral, and inferior coverage
also removes useful context. Second, the local
ROI stage was not reliable under the Vectra domain shift. Stage~2 improved the
CT MediaPipe-bbox input from 5.01 to 4.19~mm, but worsened the Vectra input
from 7.94 to 8.45~mm, which is consistent with an unresolved local
domain/correspondence mismatch.

We further asked whether the residual apparent Vectra error contained a
configuration-level pose or scale component. For each scan, we fitted an
analysis-only transform from the 24 predicted landmarks to the 24 reference
landmarks and recomputed the residual error. This uses reference landmarks and
is therefore not a deployable registration method. It is only a diagnostic for
how much of the apparent optical-to-CT error can be explained by global
configuration misalignment after prediction.

\begin{table}[t]
\caption{Analysis-only landmark-configuration alignment for Vectra predictions.
The transform is fitted from predicted landmarks to reference landmarks for
each scan and is not deployable. Values are landmark-instance MREs in
millimeters over the 14 Vectra scans. Trans. denotes translation-only alignment,
Rigid denotes rotation and translation, and Sim. denotes similarity alignment.}
\label{tab:supp_vectra_similarity_alignment}
\centering
\scriptsize
\setlength{\tabcolsep}{3.6pt}
\begin{tabular}{lcccccc}
\toprule
Condition & Raw & Trans. & Rigid & Sim. & Sim. SDR@5 & Sim. Fail@10\\
 & (mm) & (mm) & (mm) & (mm) & (\%) & (\%)\\
\midrule
Original Stage~1 & 12.40 & 10.28 & 7.64 & 7.53 & 29.8 & 22.0\\
Original final ShapeGate & 10.86 & 9.10 & 6.19 & 5.55 & 50.0 & 9.2\\
MediaPipe-bbox Stage~1 & 7.94 & 6.46 & 4.30 & 4.07 & 69.3 & 1.5\\
MediaPipe-bbox Stage~2 & 8.45 & 7.11 & 4.64 & 4.28 & 69.3 & 2.1\\
\bottomrule
\end{tabular}
\end{table}

The alignment diagnostic reveals a large global landmark-configuration
component consistent with residual pose or registration mismatch, rather than
only per-landmark surface inference error. For the MediaPipe-bbox Stage~1
model, the analysis-only similarity-aligned residual MRE was 4.07~mm, with
Fail@10 reduced from 28.3\% to 1.5\%. Translation-only alignment explained a
smaller part of the error (7.94 to 6.46~mm), while rigid alignment accounted
for most of the reduction (7.94 to 4.30~mm), with scale adding only a small
additional gain. The same pattern held for the original final model, for which
the analysis-only similarity-aligned residual was 5.55~mm. Because this
alignment uses the reference landmarks, it should be read as an
error-decomposition diagnostic rather than as evidence of deployable optical
performance.

\begin{table}[t]
\caption{Post-hoc surface-correspondence diagnostics for the mesh-sampled MediaPipe-bbox Vectra inputs. Nearest-neighbor (NN) distances compare each Vectra scan to the paired CT-derived MediaPipe-bbox surface. Correlations use 14 scans from eight patients, do not treat repeated scans as independent patient samples, and are descriptive.}
\label{tab:supp_vectra_surface_distance}
\centering
\scriptsize
\setlength{\tabcolsep}{4.0pt}
\begin{tabular}{lc}
\toprule
Diagnostic quantity & Value\\
\midrule
Vectra-to-CT median NN distance, mean (range) & 3.52~mm (1.88--6.75)\\
Vectra-to-CT P95 NN distance, mean & 19.56~mm\\
Symmetric median NN distance, mean & 3.68~mm\\
Corr. median NN distance with Stage-1 G0 MRE & $r=0.75$\\
Corr. median NN distance with Stage-1 G1 MRE & $r=0.70$\\
Corr. median NN distance with Stage-2 G0 MRE & $r=0.79$\\
Corr. median NN distance with Stage-2 G1 MRE & $r=0.75$\\
\bottomrule
\end{tabular}
\end{table}

These surface-distance diagnostics are consistent with visual inspection: even
inside the same MediaPipe-bounding-box face region, paired CT and Vectra
surfaces differ by several millimeters for many scans. The relatively poor
G0/G1 Vectra errors should therefore not be interpreted solely as failure to
infer visible or near-surface landmarks; the visible soft-tissue surface itself
is not in one-to-one correspondence with the CT-derived reference surface. The
lower G3 error under the coverage-standardized Stage~1 model is better
interpreted as reduced sensitivity to local surface mismatch and stronger
dependence on global configuration, rather than as evidence that deep
landmarks are intrinsically easier from real optical scans.

This pilot is intentionally interpreted as a transfer diagnostic rather than external validation. It shows that a model trained on coordinate-consistent CT-derived surfaces does not directly transfer to these Vectra scans under the current preprocessing, but it also shows that the failure is not monolithic. Cropping CT-derived surfaces to the Vectra-observed region increased patient-mean MRE from 2.79 to 3.79~mm, whereas full-pipeline Vectra input reached 10.96~mm. This indicates that coverage alone is not an additive explanation. CT-only retraining with standardized coverage removed a large part of the Stage~1 domain gap, and landmark configuration alignment revealed a large global configuration component consistent with residual pose or registration mismatch. Surface proximity diagnostics support a mixed explanation: cases with poorer surface proximity to the paired CT surface tended to have larger errors, and Stage~2 worsened the retrained Vectra input with standardized coverage, suggesting that local geometry was present but not distributionally matched to the CT-derived training surfaces. A separate qualitative, analysis-only ICP attempt was unstable under partial overlap and is not treated as quantitative evidence. Future optical validation should therefore prospectively standardize expression, eye state, head posture, and fixation protocols, record CT--scan intervals, and separate scanner repeatability, registration, inter-acquisition deformation, and landmark prediction error \cite{heike2010stereophotogrammetry,ozsoy2015position,tangthaweesuk2025accuracy}.

\section{Clinical Measurement Details}
All cephalometric measurements are computed identically from predicted and reference landmarks. The Frankfort Horizontal is derived from Porion and Orbitale, and angular measurements are computed after the same sagittal-plane projection for both predicted and reference configurations. The reported sagittal classes use ANB thresholds: Class III when ANB $\leq 0^{\circ}$, Class I when $0^{\circ} < \mathrm{ANB} < 4^{\circ}$, and Class II when ANB $\geq 4^{\circ}$. Because SNA and SNB share Sella, ANB can be accurate even when the absolute cranial-base-dependent angles have larger errors; this is why the main paper reports both landmark-level and cephalometric-level metrics.

For the reported no-point-context ShapeGate, sagittal-class agreement was 30/31 for cases with reference ANB at least $1^{\circ}$ away from the nearest class boundary and 4/9 for cases within $1^{\circ}$ of a boundary. This boundary sensitivity is important because thresholded class labels can change despite modest continuous ANB error.

\section{Observer Reliability Limitation and Internal Repeat Attempt}

Formal protocol-matched intra-observer and inter-observer reliability was not
available for this submission. The model is evaluated against finalized
consensus landmark coordinates produced by one orthodontist and reviewed by an
experienced orthodontist. The second review was an adjudication and
quality-control step, not an independent repeated annotation, and therefore
cannot provide an inter-observer reliability estimate. Therefore, the reported prediction errors should be
interpreted as consensus-reference errors rather than distances to an error-free
anatomical truth or a human-performance benchmark.

As external context, Schlicher et al. reported 3D-CBCT
landmark-specific overall consistency of 0.50$\pm$0.24~mm for Sella and
2.68$\pm$1.32~mm for right Porion, while Hassan et al. reported total
precision ranging from 0.29$\pm$0.17 to 2.82$\pm$7.53~mm under their
protocols~\cite{schlicher2012consistency,hassan2013precision}. These values
are not protocol-matched benchmarks for the present dataset; they illustrate
that manual variability is landmark-dependent and can overlap with
millimeter-scale model errors. Systematic-review and repeated-identification
studies likewise report heterogeneous landmark reliability, with midsagittal
landmarks often more reliable than bilateral landmarks and lower repeatability
reported for points such as Gonion, Orbitale, Condylion, and Porion
\cite{sam2019reliability,kim2022reliability}.

After model selection, we attempted an internal same-annotator repeat annotation
of the locked-test landmarks using 3D Slicer rather than the original annotation
environment. This check was not protocol-matched: the software, visualization,
markup handling, and coordinate-export workflow differed from the original
annotation pipeline. During quality control, coordinate-frame/export
inconsistencies were detected in the attempted repeat workflow. Because these
cross-workflow effects could not be separated from landmark placement
variability, we do not report the resulting distances as observer-reliability
metrics and did not use them for training, hyperparameter selection, checkpoint
selection, label revision, case exclusion from the main model evaluation, or
model revision.

Several target landmarks also have qualitatively convention-dependent 3D
definitions. For example, Gonion is a cephalometric construction from
mandibular border planes rather than a unique 3D surface point; Porion,
Orbitale, and zygomatic-arch landmarks lie on elongated or irregular rims and
arches where anterior-posterior placement can vary; and soft-tissue Pog$'$ can
depend on the local chin contour and soft-tissue state. These qualitative
considerations motivate a future protocol-matched reliability study, but they
were not used to filter cases, revise labels, or modify model evaluation.

Future observer-reliability validation should use the same annotation software,
visualization settings, coordinate-export pipeline, blinding protocol, and
landmark definitions as the reference annotation workflow, and should report
intra- and inter-observer radial error by landmark and observability group.

\section{Known Failure Modes and Recommended Next Experiments}
The two contributing hospitals belong to the same university, and the pooled
patient-level split is neither site-held-out nor external validation. Because
site, scanner, age, sex, and BMI labels were unavailable, center-held-out and
subgroup analyses could not be performed. The current results therefore
indicate several failure modes that should be addressed before clinical
translation:
\begin{itemize}
\item A larger, prospectively standardized real-optical cohort should be collected near clinically indicated CT under neutral expression, recorded eye state, controlled occlusion, and standardized head posture and fixation protocols, without acquiring extra CT solely for the study.
\item Multiple repeated optical scans should quantify scanner repeatability and registration failure separately from model error.
\item The frozen point-level degradation stress tests do not represent structured non-rigid facial deformation from expression, posture, or inter-acquisition change; future stress tests should model these coherent changes explicitly.
\item Future re-curated or prospective datasets should retain non-identifying site, scanner, and demographic metadata to enable center-held-out, scanner-held-out, and subgroup validation.
\item Formal protocol-matched landmark reliability studies should quantify
same-workflow intra-observer and inter-observer radial error by landmark and
observability group, using the same annotation software, visualization settings,
coordinate-export pipeline, blinding protocol, and landmark definitions as the
reference annotation workflow.
\item Severe sparsity should be addressed by sparse-aware pretraining or density augmentation; gate magnitude should be calibrated against actual error and uncertainty.
\end{itemize}
\FloatBarrier

\end{document}